\documentclass[10pt,twocolumn,letterpaper]{article}

\usepackage{cvpr}              % To produce the CAMERA-READY version
\usepackage{graphicx}
\usepackage{amsmath}
\usepackage{amssymb}
\usepackage{booktabs}
\usepackage{tabularx}
\usepackage{bbold}
\usepackage{subcaption}
\usepackage{textcomp}  % Required for encoding \textbigcircle
\usepackage{scalerel}  % Required for emoji \scalerel
\usepackage{caption}
\usepackage{comment}
\usepackage{pifont}
\usepackage{wasysym}
\usepackage{graphicx}
\usepackage{bbding}

\usepackage{color}
\usepackage{colortbl}
\usepackage{multirow}

\usepackage{xparse}

\usepackage[accsupp]{axessibility}  % Improves PDF readability for those with disabilities.

\usepackage[usenames,dvipsnames]{xcolor}

\NewDocumentCommand\icon{}{\scalerel*{\includegraphics{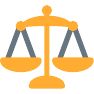}}{X}}
\newcommand{\modelname}{LIBRA\hspace{-0.8mm}{\fontsize{12pt}{12pt}\selectfont ~\icon~}}

\newcommand{\ctog}{\textcolor{Violet}{$\text{context} \rightarrow \text{gender}$}\xspace}
\newcommand{\gtoc}{\textcolor{RedOrange}{$\text{gender} \rightarrow \text{context}$}\xspace}

\definecolor{oursrow}{HTML}{E6F2F4}

\usepackage{hyperref}
\hypersetup{pagebackref,colorlinks}

\usepackage[capitalize]{cleveref}
\crefname{section}{Sec.}{Secs.}
\Crefname{section}{Section}{Sections}
\Crefname{table}{Table}{Tables}
\crefname{table}{Tab.}{Tabs.}

\def\cvprPaperID{4117} % *** Enter the CVPR Paper ID here
\def\confName{CVPR}
\def\confYear{2023}

\title{Model-Agnostic Gender Debiased Image Captioning}
\author{Yusuke Hirota \hspace{35pt} Yuta Nakashima \hspace{35pt} Noa Garcia\\
{\tt\small \{y-hirota@is., n-yuta@, noagarcia@\}ids.osaka-u.ac.jp}\\
Osaka University
}

\begin{document}

\twocolumn[{%
\renewcommand\twocolumn[1][]{#1}%
\maketitle
\begin{center}
  \centering
  \vspace{-10pt}
  \includegraphics[clip, width=0.98\textwidth]{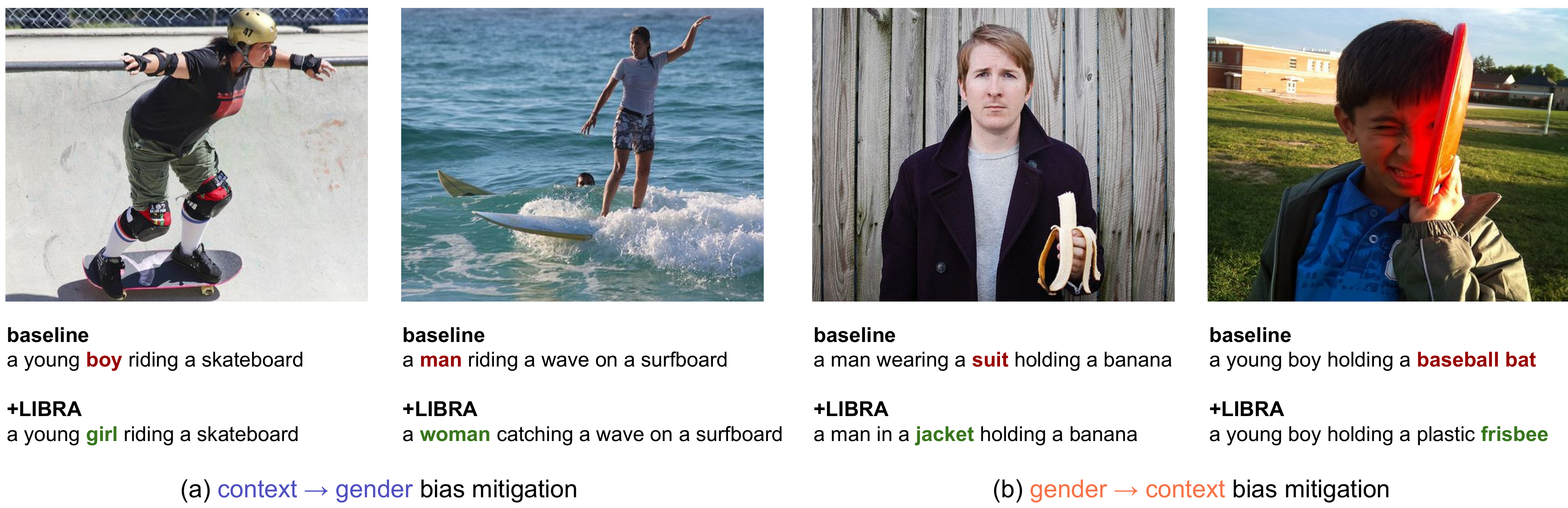}
    % \vspace{-5pt}
    \captionsetup{type=figure}
    \vspace{-2pt}
  \captionof{figure}{Generated captions by a baseline captioning model (UpDn \cite{anderson2018bottom}) and LIBRA. We show the baseline suffers from \ctog/\gtoc biases, predicting incorrect gender or incorrect word (\eg, in the left example, \textit{skateboard} highly co-occurs with men in the training set, and the baseline incorrectly predicts \textit{boy}). Our proposed framework successfully modifies those incorrect words.}
  \label{fig:first}
\end{center}%
}]

%%%\maketitle

%%%%%%%%% ABSTRACT
\begin{abstract}
Image captioning models are known to perpetuate and amplify harmful societal bias in the training set. In this work, we aim to mitigate such gender bias in image captioning models. While prior work has addressed this problem by forcing models to focus on people to reduce gender misclassification, it conversely generates gender-stereotypical words at the expense of predicting the correct gender. From this observation, we hypothesize that there are two types of gender bias affecting image captioning models: 1) bias that exploits context to predict gender, and 2) bias in the probability of generating certain (often stereotypical) words because of gender. To mitigate both types of gender biases, we propose a framework, called LIBRA, that learns from synthetically biased samples to decrease both types of biases, correcting gender misclassification and changing gender-stereotypical words to more neutral ones. Code is available at \url{https://github.com/rebnej/LIBRA} .

\end{abstract}

%%%%%%%%% BODY TEXT

\section{Introduction}
\label{sec:intro}

%Societal bias (\eg, gender bias, racial bias) has recently become a significant issue in computer vision applications. Prior studies have shown that models trained on datasets containing societal bias are known to output adverse results for minority groups such as gender and race \cite{excavating,buolamwini2018gender,zhao2017men,burns2018women,wilson2019predictive,zhao2021captionbias,hirota2022quantifying}. That is one of the major barriers, especially when adapting computer vision systems to the real world \cite{d2020data}.
In computer vision, societal bias, for which a model makes adverse judgments about specific population subgroups usually underrepresented in datasets, is increasingly concerning  \cite{buolamwini2018gender,wilson2019predictive,birhane2021multimodal,shen2022fair,de2019does,khan2021one,stock2018convnets,wang2019racial,berg2022prompt,hirota2022gender}. 
A renowned example is the work by Buolamwini and Gebru \cite{buolamwini2018gender}, which demonstrated that commercial facial recognition models predict Black women with higher error rates than White men. The existence of societal bias in datasets and models is extremely problematic as it inevitably leads to discrimination with potentially harmful consequences against people in already historically discriminated groups.

One of the computer vision tasks in which societal bias is prominent is image captioning \cite{vinyals2015show,xu2015show}, which is the task of generating a sentence describing an image. 
Notably, image captioning models not only reproduce the societal bias in the training datasets, but also 
%They also amplify societal bias; a phenomenon referred to as bias amplification \cite{hall2022systematic,choi2020fair,leino2018feature,srinivasan2021worst}. Bias amplification occurs when captioning models produce more biased sentences than those in the training datasets regarding attributes like gender, increasing harmful stereotypes about that attributes.
amplify it. This phenomenon is known as bias amplification \cite{hall2022systematic,choi2020fair,leino2018feature,srinivasan2021worst,zhao2022men} and makes models produce sentences more biased than the ones in the original training dataset. As a result, the generated sentences can contain stereotypical words about attributes such as gender that are sometimes irrelevant to the images. 
%For instance, Burns \etal. \cite{burns2018women} showed that image captioning models tend to exploit superficial statistical bias in training data and predict incorrect gender words without looking at people in images. 
%Image captioning is the task of generating a caption that describes the content of a given image and is one of the main vision-and-language tasks requiring both image and language understanding. Various image captioning models have been proposed in recent years to generate more accurate captions \cite{vinyals2015show,xu2015show,you2016image,rennie2017self,anderson2018bottom,vaswani2017attention,li2020oscar}. Nevertheless, on the other hand, it has become apparent that image captioning models tend to produce captions that contain societal bias, amplifying such as gender bias \cite{burns2018women,zhao2021captionbias,hirota2022quantifying}.  This phenomenon occurs when models amplify the societal bias in datasets, referred to as bias amplification \cite{zhao2017men,burns2018women,wang2019balanced,hirota2022quantifying}.

%\begin{figure*}[t]
%  \centering
%  \includegraphics[clip, width=0.98\textwidth]{cvpr2023-author_kit-v1_1-1/latex/figures/Two_bias (9).pdf}
%  \caption{}
%  \label{fig:two_bias}
%\end{figure*}

\begin{figure*}[t]
  \centering
  \includegraphics[clip, width=0.95\textwidth]{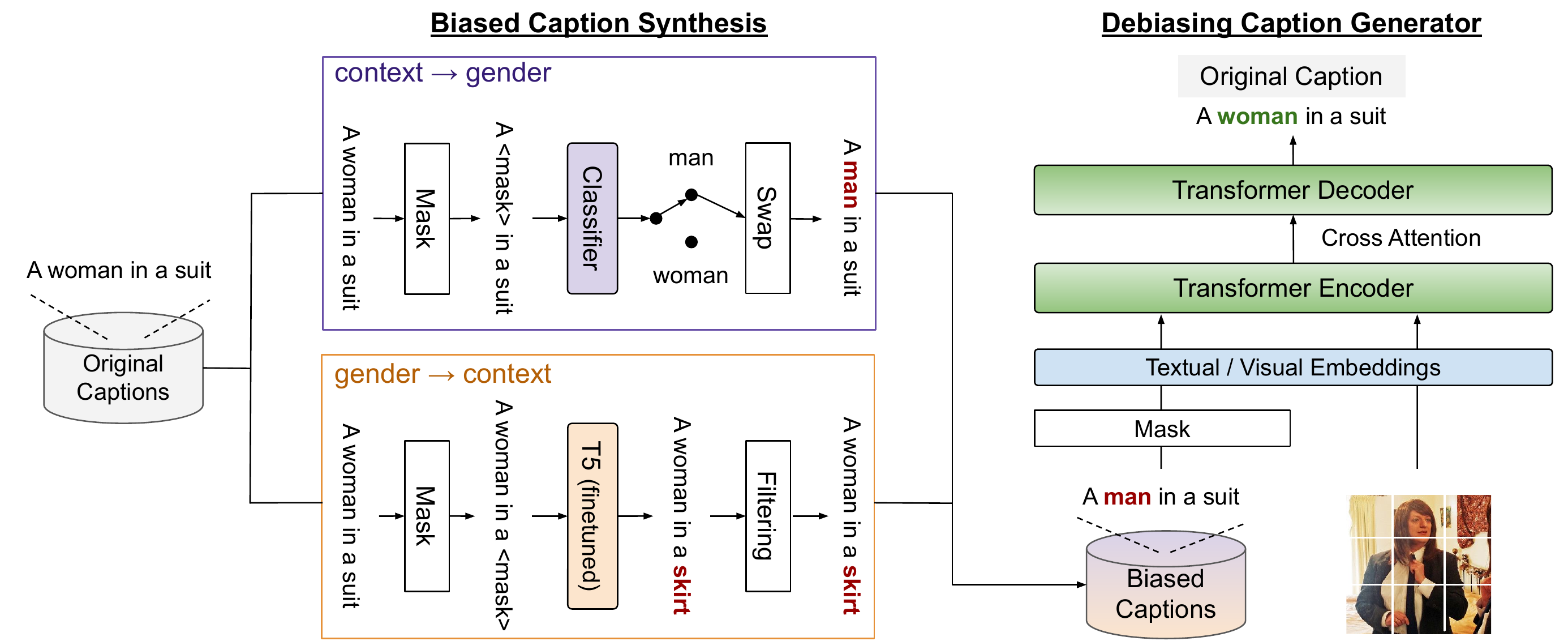}
  \caption{Overview of LIBRA. For the original captions (\ie, ground-truth captions written by annotators), we synthesize biased captions with \ctog or/and \gtoc bias (Biased Caption Synthesis). Then, given the biased captions and the original images, we train an encoder-decoder captioner, Debiasing Caption Generator, to debias the input biased captions  (\ie, predict original captions). }
  \label{fig:libra}
\end{figure*}

Our study focuses on gender bias in image captioning models. 
%Previous works have shown two types of gender bias amplification in image captioning models \cite{burns2018women,wang2021directional,hirota2022quantifying}
First, based on the observations in previous work\cite{burns2018women,wang2021directional,hirota2022quantifying,tang2021mitigating,bhargava2019exposing}, we hypothesize that there exist two different types of biases affecting captioning models:

\begin{description}
\item[\textbf{\textit{Type 1}.}] \ctog bias, which makes captioning models exploit the context of an image and precedently generated words, increasing the probability of predicting certain gender, as shown in Figure \ref{fig:first} (a).%, and a scheme of how this bias works is shown in Figure \ref{fig:two_bias} (a). 
%A captioning model exploits the context of an image and a precedently generated sentence and incorrectly predicts the gender (\ctog bias amplification) (Figure \ref{fig:first} (a), \ref{}).
%\end{quote}

%\begin{quote}
\item[\textbf{\textit{Type 2}.}] \gtoc bias, which increases the probability of generating certain words given the gender of people in an image, as shown in Figure \ref{fig:first} (b).%, and a scheme of how this bias works is shown in Figure \ref{fig:two_bias} (b).
\end{description}

\noindent
Both types of biases can result in captioning models generating harmful gender-stereotypical sentences.

A seminal method to mitigate gender bias in image captioning is Gender equalizer \cite{burns2018women}, which forces the model to focus on image regions with a person to predict their gender correctly. Training a captioning model using Gender equalizer successfully reduces gender misclassification (reducing \ctog bias).
However, focusing only on decreasing such bias can conversely amplify the other type of bias \cite{wang2021directional,hirota2022quantifying}. For example, as shown in Figure \ref{fig:equalizer}, a model trained to correctly predict the gender of a person can produce other words that are biased toward that gender (amplifying \gtoc bias). This suggests that methods for mitigating bias in captioning models must consider both types of biases.

We propose a method called \modelname (mode\underline{l}-agnost\underline{i}c de\underline{b}iasing f\underline{ra}mework) to mitigate bias amplification in image captioning by considering both types of biases. Specifically, LIBRA consists of two main modules: 1) Biased Caption Synthesis (BCS), which synthesizes gender-biased captions (Section \ref{sec:data_aug}), and 2) Debiasing Caption Generator (DCG), which mitigates bias from synthesized captions (Section \ref{sec:dcg}).
Given captions written by annotators, BCS synthesizes biased captions with \gtoc or/and \ctog biases. 
DCG is then trained to recover the original caption given a $\langle \text{synthetic biased caption, image} \rangle$ pair. 
%When training \modelname, given (synthetic biased caption, image) pairs, it is trained to recover the original caption. 
Once trained, DCG can be used on top of any image captioning models to mitigate gender bias amplification by taking the image and generated caption as input. % and producing debiased captions.
%To mitigate bias amplification in image captioning models, the trained \modelname takes generated captions from a captioning model and corresponding images and then generates debiased captions. 
Our framework is model-agnostic and does not require retraining image captioning models. % for mitigating gender bias amplification.

Extensive experiments and analysis, including quantitative and qualitative results, show that LIBRA reduces both types of gender biases in most image captioning models on various metrics
%, improving bias amplification in the metrics proposed for measuring bias amplification 
\cite{hirota2022quantifying,burns2018women,zhao2017men,tang2021mitigating}. 
This means that DCG can correct gender misclassification caused by the context of the image/words that is biased toward a certain gender, mitigating \ctog bias (Figure \ref{fig:first} (a)). Also, it tends to change words skewed toward each gender to less biased ones, mitigating \gtoc bias (Figure \ref{fig:first} (b)). Furthermore, we show that evaluation of the generated captions' quality by a metric that requires human-written captions as ground-truth (\eg, BLEU \cite{papineni2002bleu} and SPICE \cite{anderson2016spice}) likely values captions that imitate how annotators tend to describe the gender (\eg., \textit{women posing} vs.~\textit{men standing}).%, indicating the gender bias from annotators. 

\section{Related work}
\label{sec:relatedwork}

\noindent
\textbf{Societal bias in image captioning}
In image captioning \cite{you2016image,anderson2018bottom}, societal bias can come from both the visual and linguistic modalities \cite{burns2018women,zhao2021captionbias,wang2021directional}. In the visual modality, the image datasets used to train captioning models are skewed regarding human attributes such as gender \cite{zhao2017men,zhao2021captionbias,yang2020towards,ross2020measuring,garciauncurated}, in which the number of images with men is twice as much as those of women in MSCOCO \cite{lin2014microsoft}. Additionally, captions written by annotators can also be biased toward a certain gender because of gender-stereotypical expressions \cite{zhao2021captionbias,bhargava2019exposing}, which can be a source of bias from the linguistic modality. 
Models trained on such datasets not only reproduce societal bias but amplify it \cite{burns2018women,wang2021directional,zhao2021captionbias,hirota2022quantifying}. This phenomenon is demonstrated by Burns \etal. \cite{burns2018women}, which showed that image captioning models learn the association between gender and objects and make gender distribution in the predictions more skewed than in datasets. %Recently, Zhao \etal. \cite{zhao2021captionbias} investigated gender and racial bias in image captioning and found that captioning models amplify racial bias by generating stereotypical captions in terms of skin-tone. 
We show that LIBRA can mitigate such gender bias amplification in various captioning models. What is better, we demonstrate that our model often produces less gender-stereotypical captions than the original captions.

%\vspace{5pt}
%\noindent
%\textbf{Bias measurement for image captioning}
%Prior work has proposed metrics to measure societal bias in image captioning from multiple aspects \cite{burns2018women,zhao2021captionbias,hirota2022quantifying,tang2021mitigating,tang2021mitigating}. The seminal work of Burns \etal. \cite{burns2018women} measures gender misclassification rates in image captioning models, showing that models tend to predict gender words by exploiting contextual cues other than people. In the recent work of Hirota \etal. \cite{hirota2022quantifying}, bias amplification of captioning models is evaluated by measuring how much the annotator-written captions and the model-generated captions differed by gender. Besides, bias amplification measurements based on gender-object co-occurrence for image classification \cite{zhao2017men,wang2021directional,tang2021mitigating} can be extended to image captioning by considering word-gender co-occurrence, which we also show our model can mitigate gender bias amplification in these metrics. 

\vspace{5pt}
\noindent
\textbf{Mitigating societal bias}
 Mitigation of societal bias has been studied in many tasks \cite{zhao2017men,wang2019balanced,tang2021mitigating,jain2020imperfect,jia2020mitigating,zhang2022counterfactually,whitehead2022reliable,wang2020towards,thong2021feature,wang2020revise,yao2022improving}, such as image classification \cite{rawat2017deep} and visual semantic role labeling \cite{yatskar2016situation}. For example, Wang \etal. \cite{wang2019balanced} proposed an adversarial debiasing method to mitigate gender bias amplification in image classification models. In image captioning, Burns \etal \cite{burns2018women} proposed the Gender equalizer we described in Section \ref{sec:intro} to mitigate \ctog bias. 
 However, recent work \cite{wang2021directional,hirota2022quantifying} showed that focusing on mitigating gender misclassification can lead to generating gender-stereotypical words and amplifying \gtoc bias. 
 LIBRA is designed to mitigate bias from the two types of biases. 
 %1) Biased Caption Synthesis (BCS) for synthesizing biased captions that contain two types of bias and 2) a model-agnostic debiasing generator (\modelname) that takes (caption, image) pairs and generates debiased captions. 

\vspace{5pt}
\noindent
\textbf{Image caption editing}
DCG takes a $\langle \text{caption, image} \rangle$ pair as input and debiases the caption. This process is aligned with image caption editing \cite{sammani2020show,sammani2019look,wang2022explicit} for generating a refined caption. These models aim to correct grammatical errors and unnatural sentences but not to mitigate gender bias.  
In Section \ref{sec:edit}, we compare DCG with a state-of-the-art image caption editing model \cite{sammani2020show} and show that a dedicated framework for addressing gender bias is necessary.

\section{Biased caption synthesis}
\label{sec:data_aug}

Figure \ref{fig:libra} shows an overview of LIBRA, consisting of BCS and DCG. This section introduces BCS to synthesize captions with both \ctog or/and \gtoc biases. 

\vspace{5pt}
\noindent
\textbf{Notation} Let $\mathcal{D} = \{(I, y)\}$ denote a training set of the captioning dataset, where $I$ is an image and $y = (y_1, \dots, y_N)$ is the ground-truth caption with $N$ tokens. $\mathcal{D}_\text{g}$ denotes a subset of $\mathcal{D}$,
which is given by filter $F_\text{GW}$ as
\begin{equation}
    \mathcal{D}_\text{g} = F_\text{GW}(\mathcal{D}),
\end{equation}
$F_\text{GW}$ keeps captions that contains either women or men words (\eg, \textit{girl}, \textit{boy}).\footnote{We pre-defined women and men words. The list is in the appendix.} Therefore, samples in $\mathcal{D}_\text{g}$ come with a gender attribute $g \in \mathcal{G}$, where $\mathcal{G} = \{\text{female}, \text{male}\}$.\footnote{In this paper, we focus on binary gender categories in our framework and evaluation by following previous work \cite{zhao2017men,burns2018women}. We recognize that the more inclusive gender categories are preferable, and it is the future work.} We define the set that consists of women and men words as gender words.

\subsection{\textcolor{Violet}{Context} \textcolor{Violet}{$\rightarrow$} \textcolor{Violet}{gender} bias synthesis}
\label{sec:gs}

\begin{figure}[t]
  \centering
  \includegraphics[clip, width=1.0\columnwidth]{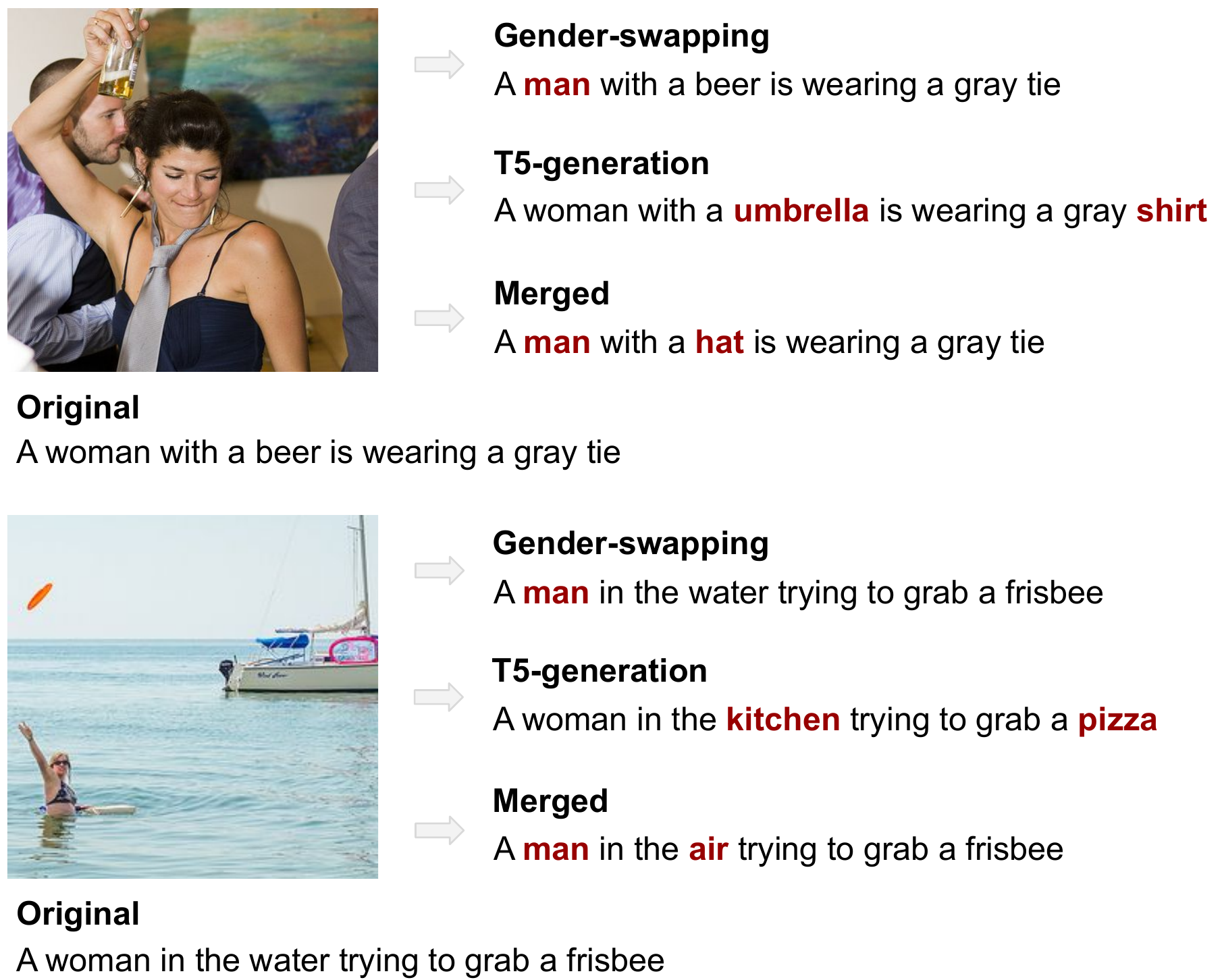}
  \caption{Biased captions synthesized by BCS. Gender-swapping denotes synthesized captions by swapping the gender words (Section \ref{sec:gs}). T5-generation denotes synthesized captions by T5 (Section \ref{sec:t5}). Merged represents biased captions synthesized by applying T5-generation and Gender-swapping (Section \ref{sec:t5_gs}).}
  \label{fig:syn_cap}
\end{figure}

\textcolor{Violet}{Context} \textcolor{Violet}{$\rightarrow$} \textcolor{Violet}{gender} bias means gender prediction is overly contextualized by the image and caption. 
Therefore, the gender should be predictable from the image and caption context when the caption has \ctog bias. 
The idea of synthesizing \ctog biased captions is thus to swap the gender words in the original caption to make it consistent with the context when the gender predicted from the context is skewed toward the other gender. Since an original caption faithfully represents the main content of the corresponding image \cite{tavakoli2017paying,berg2012understanding}, we can solely use the caption to judge if both image and caption are skewed. To this end, we train a sentence classifier that predicts gender from textual context to synthesize biased captions. We introduce the detailed steps. 

\vspace{-10pt}
\paragraph{Masking}
Captions with \ctog bias are synthesized for $\mathcal{D}_\text{g}$. Let $F_\text{PG}$ denote the filter that removes captions whose gender is predictable by the sentence classifier. 
Given $y \in \mathcal{D}_\text{g}$, $F_\text{PG}$ instantiated by first masking gender words and replacing corresponding tokens with the mask token to avoid revealing the gender, following \cite{hirota2022quantifying}. We denote this gender word masking by $\text{mask}(\cdot)$.
%Given caption $y \in \mathcal{D}_\text{g}$, we first mask gender words by replacing corresponding tokens with the mask token to avoid revealing the gender, following \cite{hirota2022quantifying}. Note that we can assume the masked caption $y'$ describes the content of an image other than gender as the gender words are masked.

\vspace{-10pt}
\paragraph{Gender classifier}
We then train\footnote{Refer to the appendix for training details.} gender classifier $f_\text{g}$ to predict the gender from masked caption as
\begin{equation}
    \hat{g} = f_\text{g}(y) = \text{argmax}_g \ p(G=g | \text{mask}(y))
\end{equation}
where $p(G=g | \text{mask}(y))$ is the probability of being gender $g$ given masked $y$. 
$F_\text{PG}$ is then applied to $\mathcal{D}_\text{g}$ as:
\begin{equation}
    F_\text{PG}(\mathcal{D}_\text{g}) = \{y \in \mathcal{D}_\text{g}|\hat{g}(y) \neq g\},\label{eq:gs1}
\end{equation}
recalling $\hat{g}$ is a function of $y$.

%We train a gender classifier $f_\text{g}$ that predicts gender $g$ from the masked captions as
%
%\begin{equation}
%    \hat{g} = f_\text{g}(y') = \text{argmax}_g \ p(G=g | y')
%\end{equation}
%where $p(G=g | y')$ is the probability of being gender $g$. The trained $f_\text{g}$ learns to predict gender from the context of images.% because $y'$ describes the content other than gender.

%To synthesize biased captions, we rely on a classifier that predicts gender from textual context. Specifically, given a caption $y \in \mathcal{D}_\text{g}$, we first mask the gender words (\ie, women or men words) to avoid revealing the gender. The masked caption $y'$ is fed into a gender classifier $f_\text{g}$ trained to predict gender $g$, which is a binary classification as
%
%\begin{equation}
%    \hat{g} = f_\text{g}(y') = \text{argmax}_g \ p(G=g | y')
%\end{equation}
%where $p(G=g | y')$ is the probability of being gender $g$. 
%The classifier is trained on the captions in $\mathcal{D}_\text{g}$. Note that we can assume $f_\text{g}$ learns to predict gender from the context of images as $y'$ describes the content other than gender. 

\vspace{-10pt}
\paragraph{Gender swapping}
The inconsistency of context $y'$ and gender $g$ means that $y'$ is skewed toward the other gender; therefore, swapping gender wards (\eg, \textit{man} $\rightarrow$ \textit{woman}) in $y \in F_\text{PG}(\mathcal{D}_\text{g})$ results in a biased caption. Letting $\text{swap}(\cdot)$ denote this gender swapping operation, the augmenting set $\mathcal{A}_\text{CG}$ is given by: 
\begin{equation}
    \mathcal{A}_\text{CG} = \{\text{swap}(y) | y \in F_\text{PG}(\mathcal{D}_\text{g})\}. \label{eq:gs2}
\end{equation}
%When synthesizing biased captions, we feed the masked caption $y'$ into the trained classifier and compare $\hat{g}$ with $g$ to see if the context of the image matches the gender in the original caption. If they do not match, the context of the image is not skewed toward the gender. Thus, in that case, we swap the gender words in $y$ into the other gender words (\eg, \textit{man} $\rightarrow$ \textit{woman}) to synthesize a caption with \ctog bias, which contains incorrect gender words due to the context of the images. %As a result, the synthesized captions contain \ctog bias because the incorrect gender in them is due to the content of the images except for people's gender.  

Figure \ref{fig:syn_cap} shows some synthetically biased captions (refer to Gender-swapping). We can see that the incorrect gender correlates with context skewed toward that gender. For instance, in the top example, \textit{tie} is skewed toward men based on the co-occurrence of men words and \textit{tie}. 
%We show some synthesized captions in Figure \ref{fig:syn_cap} (refer to Gender swapping). We can see that the incorrect gender correlates with context skewed toward that gender. For instance, in the top example, \textit{tie} is skewed toward men based on the co-occurrence of men words and \textit{tie}. 

%When synthesizing biased captions, we feed the original caption $y$ into the trained classifier and compare $\hat{g}$ with $g$ to see if the context of the image matches the gender in the original caption. If they are not matched, we swap the gender words in $y$ into the other gender words (\eg, \textit{man} $\rightarrow$ \textit{woman}).

 %The synthesized captions contain \ctog bias because the incorrect gender in the synthetic captions is due to the content of the images except for people's gender. We show the synthesized captions in Figure \ref{fig:syn_cap} (refer to GS), and we can see that the incorrect gender correlates with context skewed toward that gender (\eg, in the top example, \textit{tie} is skewed toward men based on the co-occurrence of men words and \textit{tie}). 

\subsection{\textcolor{RedOrange}{Gender} \textcolor{RedOrange}{$\rightarrow$} \textcolor{RedOrange}{context} bias synthesis}
\label{sec:t5}

Our idea for synthesizing captions with \gtoc bias is to sample randomly modified captions of $y$ and keep ones with the bias. Sampling modified captions that potentially suffer from this type of bias is not trivial. We thus borrow the power of a language model. That is, captions with \gtoc bias tend to contain words that well co-occur with gender words, and this tendency is supposedly encapsulated in a language model trained with a large-scale text corpus. We propose to use the masked token generation capability of T5 \cite{raffel2020exploring} to sample modified captions and filter them for selecting biased captions. 
%For synthesizing such biased captions, which contain words skewed toward the gender in images, we propose masked word generation with T5 \cite{raffel2020exploring} and filters for selecting high-quality and biased captions. 

%For synthesizing biased captions, we rely on masked word generation by T5 \cite{raffel2020exploring}\alert{, a Transformer language model based on a text-to-text approach}. 
%The process of synthesizing biased captions is shown in Figure \ref{fig:t5}.

%\paragraph{Finetuning T5}
%Following the masked language model in \cite{devlin2018bert}, we finetune T5, a Transformer language model based on a text-to-text approach, on captions in $\mathcal{D}$ to make the language model learn the vocabulary in the captioning dataset. Specifically, we mask $10$\% of the words in the original caption $y$. Given the masked caption and the positions of masked words $\mathcal{M} = \{m_1, \dots ,m_{|\mathcal{M}|}\}$, T5 predicts the probability of masked words by $\prod_{m \in \mathcal{M}} p(y_m|y_{\backslash \mathcal{M}})$, where $y_{\backslash \mathcal{M}}$ denotes all words in an input caption $y$ except for masked words $y_m$. The loss of this finetuning task is defined as: 

%\begin{equation}
%    \mathcal{L}_{mlm} = -\sum_{w \in \mathcal{W}} \text{log} \prod_{m \in \mathcal{M}} p(y_m|y_{\backslash \mathcal{M}};\Theta)
%\end{equation}
%where $\mathcal{W}$ is the pre-defined vocabulary of T5.  

\vspace{-10pt}
\paragraph{T5 masked word generation}
T5 is one of the state-of-the-art Transformer language models. For better alignment with the vocabulary in the captioning dataset, we finetune T5 with $\mathcal{D}$ by following the process of training the masked language model in \cite{devlin2018bert}.\footnote{Refer to the appendix for the details of this finetuning.}
After finetuning, we sample randomly modified captions using T5. Specifically, we randomly mask $15$\% of the tokens in $y \in \mathcal{D}$. %to decide the tokens to be changed by T5.
Note that we exclude tokens of the gender words if any as they serve as the only cue of the directionality of bias (either men or women). 

Let $y_\mathcal{M}$ denote a modified $y$ whose $m$-th token ($m \in \mathcal{M}$) is replaced with the mask token. The masked token generator by T5 can complete the masked tokens solely based on $y_\mathcal{M}$, \ie, $\hat{y} = \text{T5}(y_\mathcal{M})$. With this, we can sample an arbitrary number of $\hat{y}$'s to make a T5-augmented set $\mathcal{D}_\text{T5}$ as\footnote{We remove trivial modification that replaces a word with its synonyms based on WordNet \cite{miller1995wordnet} and unnatural captions with dedicated classifier. More details can be found in the appendix.}:
\begin{equation}
    \mathcal{D}_\text{T5} = \{\hat{y} = \text{T5}(y_\mathcal{M})| y \in \mathcal{D}, \mathcal{M} \sim \mathcal{R}\},
\end{equation}
where $\mathcal{M}$ is sampled from set $\mathcal{R}$ of all possible masks. 
%After finetuning, we synthesize biased captions by using T5. Specifically, we first mask $15$\% of the words in the input caption. We do not mask gender-related words as we focus on \gtoc bias, which intends to generate gender-stereotypical words based on the gender in images. Given the masked captions, T5 predicts the words that suit the masked positions according to the language model learned during the fine-tuning phase. We do not generate synonyms of masked words to avoid trivial changes by using WordNet \cite{miller1995wordnet}. More details about masked word generation can be found in the appendix.

%\begin{equation}
%   w_{m}^\star = \text{argmax}_{w \in \mathcal{W}} \ p(w|y_{\backslash M};\Theta) \ \ \text{for all} \ m \in M
%\end{equation}
%where $w_{m}^\star$ 

%\vspace{-10pt}
\paragraph{Filtering}
We then apply a filter to $\mathcal{D}_\text{T5}$ to remove captions that decrease \gtoc bias, which is referred to as gender filter. 
We thus borrow the idea in Eq.~(\ref{eq:gs1}). For this, we only use captions in $\mathcal{D}_\text{T5}$ that contain the gender words, \ie, $\mathcal{D}_\text{T5,g} = F_\text{GW}(\mathcal{D}_\text{T5})$, to guarantee that all captions have gender attribute $g$. To collectively increase \gtoc bias in the set, we additionally use condition $d(y',y) = p(G=g | \text{mask}(y')) -  p(G=g | \text{mask}(y)) > \delta $, which means the gender of $y' \in \mathcal{D}_\text{T5,g}$ should be more predictable than the corresponding original $y \in \mathcal{D}_\text{g}$ by a predefined margin $\delta$. Gender filter $F_\text{GF}$ is given by:
\begin{equation}
    F_\text{GF}(\mathcal{D}_\text{T5,g}, \mathcal{D}_\text{g}) = \{y' \in \mathcal{D}_\text{T5,g}|\hat{g}(y') = g, d(y', y) > \delta\}.
\end{equation}
The appendix shows that $F_\text{GF}$ can keep more gender-stereotypical sentences than the original captions.

With the gender filter, augmenting set $\mathcal{A}_\text{GC}$ is given as the intersection of the filtered sets as:
\begin{equation}
    \mathcal{A}_\text{GC} = F_\text{GF}(\mathcal{D}_\text{T5,g}, \mathcal{D}_\text{g}) .
\end{equation}

 As a result, the synthesized captions contain gender-stereotypical words that often co-occur with that gender as shown in Figure \ref{fig:syn_cap} (refer to T5-generation). For example, in the bottom sample, \textit{kitchen} co-occurs with women words about twice as often as it co-occurs with men words in $\mathcal{D}$, amplifying \gtoc bias.

\subsection{Merging together} 
\label{sec:t5_gs}
For further augmenting captions, we merge the processes for augmenting both \ctog and \gtoc biases, which is given by:
\begin{equation}
    \mathcal{A} = \{\text{swap}(y) | y \in F_\text{PG}(\mathcal{D}_\text{T5,g})\},
\end{equation}
which means that the process for synthesizing \ctog bias in Eqs.~(\ref{eq:gs1}) and (\ref{eq:gs2}) is applied to T5 augmented captions. 
In this way, the textual context becomes skewed toward the new gender. Some synthesized samples can be found in Figure \ref{fig:syn_cap} (refer to Merged).

\section{Debiasing caption generator}
\label{sec:dcg}

DCG is designed to mitigate the two types of gender bias in an input caption to generate a debiased caption.

\vspace{5pt}
\noindent
\textbf{Architecture} 
DCG has an encoder-decoder architecture. The encoder is a Transformer-based vision-and-language model \cite{kim2021vilt} that takes an image and text as input and outputs a multi-modal representation. The decoder is a Transformer-based language model \cite{radford2019language} that generates text given the encoder's output. The encoder's output is fed into the decoder via a cross-attention mechanism \cite{rothe2020leveraging}.   

\vspace{5pt}
\noindent
\textbf{Training} 
Let $\mathcal{D}^\star = \mathcal{A}_\text{CG} \cup \mathcal{A}_\text{GC} \cup \mathcal{A} = \{(I, y^\star)\}$ denote the set of synthetic biased captions where $y^\star$ is a biased caption.  
When training DCG, given a ($I$, $y^\star$) pair, we first mask $100  \eta$ percent of words in the input caption. The aim is to add noise to the input sentence so DCG can see the image when refining the input caption, avoiding outputting the input sentence as it is by ignoring the image. 
The masked caption is embedded to $\bar{y}$ by word embedding and position embedding. The input image $I$ is embedded to $\bar{I}$ through linear projection and position embedding.
$\bar{y}$ and $\bar{I}$ are fed into the DCG encoder, and the output representation of the encoder is inputted to the DCG decoder via a cross-attention mechanism. DCG is trained to recover the original caption $y$ with a cross-entropy loss $\mathcal{L}_\text{ce}$ as 

\begin{equation}
    \mathcal{L}_\text{ce} = -\sum_{t=1}^{N} \text{log} \  p(y_t|y_{1:t-1}, I, y^\star) 
\end{equation}
where $p$ is conditioned on the precedently generated tokens, and $I$ and $y^\star$ through the cross-attention from the encoder. The trained DCG learns to mitigate two types of biases that lie in the input-biased captions. 

\vspace{5pt}
\noindent
\textbf{Inference} 
We apply the trained DCG to the output captions of captioning models. Let $y_\text{c}$ denote a generated caption by an image captioning model. As in training, given a pair of ($I$, $y_\text{c}$), we first mask $100 \eta$ percent of words in the input caption. Then, DCG takes the masked caption and image and generates a debiased caption. DCG can be used on top of any image captioning models and does not require training in captioning models to mitigate gender bias.

\section{Experiments}
\label{sec:evaluation}

\begin{table}[t]
\renewcommand{\arraystretch}{1.1}
\setlength{\tabcolsep}{5pt}
\small
\centering
\caption{Dataset construction. Swap denotes synthesized captions by Gender-swapping (Section \ref{sec:gs}). T5 denotes synthesized captions by T5-generation (Section \ref{sec:t5}). Ratio represents the ratio of the number of each type of biased data.}
\begin{tabularx}{0.346\textwidth}{c c c c r}
\toprule
\multicolumn{3}{c}{Synthesis method}\\ \cline{0-2}  
Swap & T5 & Merged & Ratio & Num. sample \\
\midrule
\Checkmark & \Checkmark  & - & 1:1:0 & 57,284\\
- & \Checkmark & \Checkmark  & 0:1:1 & 114,568\\
 \Checkmark & \Checkmark & \Checkmark  & 1:2:1 & 114,568\\
\bottomrule
\end{tabularx}
\label{tab:dataset}
\end{table}

\definecolor{colbest}{rgb}{0.1, 0.6, 0.1}\
\definecolor{colworst}{rgb}{0.75, 0, 0}

\begin{table*}[t]
\renewcommand{\arraystretch}{1.1}
\setlength{\tabcolsep}{5pt}
\small
\centering
\caption{Gender bias and captioning quality for several image captioning models. \textcolor{colbest}{Green}/\textcolor{colworst}{red} denotes LIBRA mitigates/amplifies bias with respect to the baselines. For bias, lower is better. For captioning quality, higher is better. LIC and BiasAmp are scaled by $100$. Note that CLIPScore for ClipCap can be higher because CLIPScore and ClipCap use CLIP \cite{radford2021learning} in their frameworks.}
\begin{tabularx}{0.78\textwidth}{l r r r c r r r r r}
\toprule
& \multicolumn{3}{c}{Gender bias $\downarrow$} & & \multicolumn{5}{c}{Captioning quality $\uparrow$} \\ \cline{2-4} \cline{6-10}
Model & LIC   & Error & BiasAmp & & BLEU-4	& CIDEr & METEOR & SPICE & CLIPScore \\
\midrule
\cellcolor{oursrow} NIC \cite{vinyals2015show} & \cellcolor{oursrow} 0.5  & \cellcolor{oursrow} 23.6 & \cellcolor{oursrow} 1.61 & \cellcolor{oursrow} & \cellcolor{oursrow} 21.9 & \cellcolor{oursrow} 58.3 & \cellcolor{oursrow} 21.6 & \cellcolor{oursrow} 13.4 & \cellcolor{oursrow} 65.2\\
\hspace{2pt} +LIBRA  & \textcolor{colbest}{-0.3}  & \textcolor{colbest}{5.7} & \textcolor{colbest}{-1.47} && 24.6 & 72.0 & 24.2 & 16.5 & 71.7\\
\cellcolor{oursrow} SAT \cite{xu2015show} & \cellcolor{oursrow} -0.3  & \cellcolor{oursrow} 9.1 & \cellcolor{oursrow} 0.92 & \cellcolor{oursrow} & \cellcolor{oursrow} 34.5 & \cellcolor{oursrow} 94.6 & \cellcolor{oursrow} 27.3 & \cellcolor{oursrow} 19.2 & \cellcolor{oursrow} 72.1\\
\hspace{2pt} +LIBRA  & \textcolor{colbest}{-1.4}  & \textcolor{colbest}{3.9} & \textcolor{colbest}{-0.48} && 34.6 & 95.9 & 27.8 & 20.0 & 73.6\\
\cellcolor{oursrow} FC \cite{rennie2017self} & \cellcolor{oursrow} 2.9  & \cellcolor{oursrow} 10.3 & \cellcolor{oursrow} 3.97 & \cellcolor{oursrow}  & \cellcolor{oursrow} 32.2 & \cellcolor{oursrow} 94.2 & \cellcolor{oursrow} 26.1 & \cellcolor{oursrow} 18.3 & \cellcolor{oursrow} 70.0\\
\hspace{2pt} +LIBRA  & \textcolor{colbest}{-0.2}  & \textcolor{colbest}{4.3} & \textcolor{colbest}{-1.11} && 32.8 & 95.9 & 27.3 & 19.7 & 72.9\\
\cellcolor{oursrow} Att2in \cite{rennie2017self} & \cellcolor{oursrow} 1.1  & \cellcolor{oursrow} 5.4 & \cellcolor{oursrow} -1.01 & \cellcolor{oursrow} & \cellcolor{oursrow} 36.7 & \cellcolor{oursrow} 102.8 & \cellcolor{oursrow} 28.4 & \cellcolor{oursrow} 20.2 & \cellcolor{oursrow} 72.6\\
\hspace{2pt} +LIBRA  & \textcolor{colbest}{-0.3} & \textcolor{colbest}{4.6} & \textcolor{colbest}{-3.39} && 35.9 & 101.7 & 28.5 & 20.6 & 73.8\\
\cellcolor{oursrow} UpDn \cite{anderson2018bottom} & \cellcolor{oursrow} 4.7 & \cellcolor{oursrow} 5.6 & \cellcolor{oursrow} 1.46 & \cellcolor{oursrow} & \cellcolor{oursrow} 39.4 & \cellcolor{oursrow} 115.1 & \cellcolor{oursrow} 29.8 & \cellcolor{oursrow} 22.0 & \cellcolor{oursrow} 73.8\\
\hspace{2pt} +LIBRA  & \textcolor{colbest}{1.5} & \textcolor{colbest}{4.5} & \textcolor{colbest}{-2.23} && 37.7 & 110.1 & 29.6 & 22.0 & 74.6\\
\cellcolor{oursrow} Transformer \cite{vaswani2017attention} & \cellcolor{oursrow} 5.4 & \cellcolor{oursrow} 6.9 & \cellcolor{oursrow} 0.09 & \cellcolor{oursrow}  & \cellcolor{oursrow} 35.0 & \cellcolor{oursrow} 101.5 & \cellcolor{oursrow} 28.9 & \cellcolor{oursrow} 21.1 & \cellcolor{oursrow} 75.3\\
\hspace{2pt} +LIBRA  & \textcolor{colbest}{2.3} & \textcolor{colbest}{5.0} & \textcolor{colbest}{-0.26} && 33.9 & 98.7 & 28.6 & 20.9 & 75.7\\
\cellcolor{oursrow} OSCAR \cite{li2020oscar} & \cellcolor{oursrow} 2.4  & \cellcolor{oursrow} 3.0 & \cellcolor{oursrow} 1.78 & \cellcolor{oursrow} & \cellcolor{oursrow} 39.4 & \cellcolor{oursrow} 119.8 & \cellcolor{oursrow} 32.1 & \cellcolor{oursrow} 24.0 & \cellcolor{oursrow} 75.8\\
\hspace{2pt} +LIBRA  & \textcolor{colbest}{0.3} & \textcolor{colworst}{4.6} & \textcolor{colbest}{-1.95} && 37.2 & 113.1 & 31.1 & 23.2 & 75.7\\
\cellcolor{oursrow} ClipCap \cite{mokady2021clipcap} & \cellcolor{oursrow} 1.1 & \cellcolor{oursrow} 5.6 & \cellcolor{oursrow} 1.51 & \cellcolor{oursrow} & \cellcolor{oursrow} 34.8 & \cellcolor{oursrow} 103.7 & \cellcolor{oursrow} 29.6 & \cellcolor{oursrow} 21.5 & \cellcolor{oursrow} 76.6\\
\hspace{2pt} +LIBRA  & \textcolor{colbest}{-1.5} & \textcolor{colbest}{4.5} & \textcolor{colbest}{-0.57} && 33.8 & 100.6 & 29.3 & 21.4 & 76.0\\
\cellcolor{oursrow} GRIT \cite{nguyen2022grit} & \cellcolor{oursrow} 3.1  & \cellcolor{oursrow} 3.5 & \cellcolor{oursrow} 3.05 & \cellcolor{oursrow} & \cellcolor{oursrow} 42.9 & \cellcolor{oursrow} 123.3 & \cellcolor{oursrow} 31.5 & \cellcolor{oursrow} 23.4 & \cellcolor{oursrow} 76.2\\
\hspace{2pt} +LIBRA  & \textcolor{colbest}{0.7} & \textcolor{colworst}{4.1} & \textcolor{colbest}{1.57} && 40.5 & 116.8 & 30.6 & 22.6 & 75.9\\
\bottomrule
\end{tabularx}
\label{tab:main}
\end{table*}

\vspace{-5mm}
\paragraph{Dataset}
We use MSCOCO captions \cite{chen2015microsoft}. % as an image captioning dataset. 
For training captioning models, we use the MSCOCO training set that contains $82,783$ images. For evaluation, we use a subset of the MSCOCO validation set, consisting of $10,780$ images, that come with binary gender annotations from \cite{zhao2021captionbias}. Each image has five captions from annotators.

For synthesizing biased captions with BCS, we use the MSCOCO training set.
The maximum number of synthetic captions by Gender-swapping is capped by $|F_\text{PG}(\mathcal{D}_g)| = 28,642$, while T5-generation and Merged can synthesize an arbitrary number of captions by sampling $\mathcal{M}$. We synthesize captions so that the number of captions with gender swapping (\ie, Gender-Swapping and Merged) and T5-generation can be balanced as in \Cref{tab:dataset}. %For the second combination, we used the same number as the third combination, although T5-generation and Merged are uncapped. 

%When training DCG, we construct the biased dataset to ensure that the ratio of the biased samples with gender swapping to those without gender swapping is the same. We show the possible combinations of biased data in Table \ref{tab:dataset}.
%The maximum number of captions generated by Gender-swapping is limited to the number of captions in $\mathcal{D}_g$ which is $28,642$. The number of samples of the dataset combinations is decided by following the ratio as shown in Table \ref{tab:dataset}. In the experiments (Section \ref{sec:ablation}), we compare these combinations.

\vspace{-3mm}
\paragraph{Bias metrics}
We mainly rely on three metrics to evaluate our framework: 1) \textbf{LIC} \cite{hirota2022quantifying}, which compares two gender classifiers' accuracies trained on either generated captions by a captioning model or human-written captions. Higher accuracy of the classifier trained on a model's predictions means that the model's captions contain more information to identify the gender in images, indicating \gtoc bias amplification, 2) \textbf{Error} \cite{burns2018women}, which measures the gender misclassification ratio of generated captions. We consider Error to evaluate \ctog bias whereas it does not directly measure this bias (discussed in the appendix) 
%(Section \ref{sec:limitations})
, and 3) \textbf{BiasAmp} \cite{zhao2017men}, a bias amplification measurement based on word-gender co-occurrence, which is possibly the cause of \gtoc bias. More details about these bias metrics are described in the appendix. 

\vspace{-3mm}
\paragraph{Captioning metrics}
The accuracy of generated captions is evaluated on reference-based metrics that require human-written captions to compute scores, specifically BLEU-4 \cite{papineni2002bleu}, CIDEr \cite{vedantam2015cider}, METEOR \cite{denkowski2014meteor}, and SPICE \cite{anderson2016spice}. While those metrics are widely used to evaluate captioning models, they often suffer from disagreements with human judges \cite{hessel2021clipscore}. Thus, we also use a reference-free metric, CLIPScore \cite{hessel2021clipscore}, that relies on the image-text matching ability of the pre-trained CLIP. CLIPScore has been shown to have a higher agreement with human judgment than reference-based metrics. %Thus we regard CLIPScore as the most important metric for accuracy evaluation.

\vspace{-3mm}
\paragraph{Captioning models}
We evaluate two standard types of captioning models as baselines: 1) CNN encoder-LSTM decoder models (NIC \cite{vinyals2015show}, SAT \cite{xu2015show}, FC \cite{rennie2017self}, Att2in \cite{rennie2017self}, and UpDn \cite{anderson2018bottom}) and 2) state-of-the-art Transformer-based models (Transformer \cite{vaswani2017attention}, OSCAR \cite{li2020oscar}, ClipCap \cite{mokady2021clipcap}, and GRIT \cite{nguyen2022grit}). Note that most of the publicly available pre-trained models are trained on the training set of the Karpathy split \cite{karpathy2015deep} that uses the training and validation sets of MSCOCO for training. As we use the MSCOCO validation set for our evaluation, we retrain the captioning models on the MSCOCO training set only.

\vspace{-3mm}
\paragraph{Debiasing methods}
As debiasing methods, we compare LIBRA against Gender equalizer \cite{burns2018women}.
Gender equalizer utilizes extra segmentation annotations in MSCOCO \cite{lin2014microsoft}, which are not always available.
% Gender equalizer \cite{burns2018women} is a bias mitigation method that trains a captioning model to focus on people's areas in images to predict gender correctly.
% It utilizes extra segmentation annotations in MSCOCO \cite{lin2014microsoft} to mask people in images, which are not always available. 
 The method is not applicable to captioning models that use object-based visual features such as Faster R-CNN \cite{ren2016faster} because the pre-trained detector's performance drops considerably for human-masked images.\footnote{Faster-RCNN mAP drops from $0.41$ to $0.37$, and for the person class recall drops from $0.79$ to $0.68$.}
 In the experiment, we apply Gender equalizer and LIBRA to debias NIC+, which is a variant of NIC with extra training on images of female/male presented in \cite{burns2018women}. 

For LIBRA, we use $\delta = 0.2$.
The vision-and-language encoder of DCG is Vilt \cite{kim2021vilt}, and the decoder is GPT-2 \cite{radford2019language}. 
Unless otherwise stated, we use the combination of biased data composed of T5-generation and Merged in Table \ref{tab:dataset}. We set $\eta = 0.2$ and conduct ablation studies of the settings in Section \ref{sec:ablation} and the appendix.

\subsection{Bias mitigation analysis}
\label{sec:main}

\begin{figure}[t]
  \centering
  \includegraphics[clip, width=1.0\columnwidth]{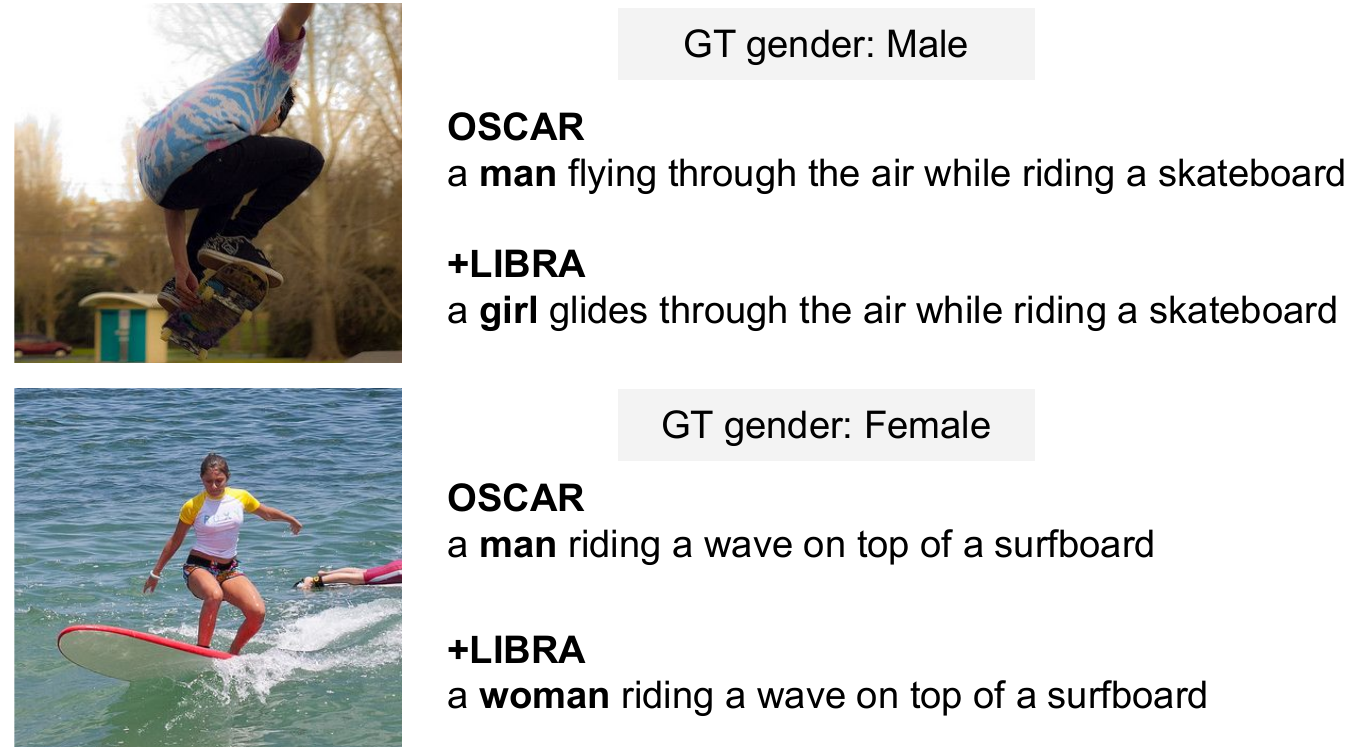}
  \caption{Gender misclassification of LIBRA (Top). Gender misclassification of OSCAR \cite{li2020oscar} (Bottom). GT gender denotes ground-truth gender annotation in \cite{zhao2021captionbias}.}
  \label{fig:miscls}
\end{figure}

\begin{figure}[t]
  \centering
  \includegraphics[clip, width=0.98\columnwidth]{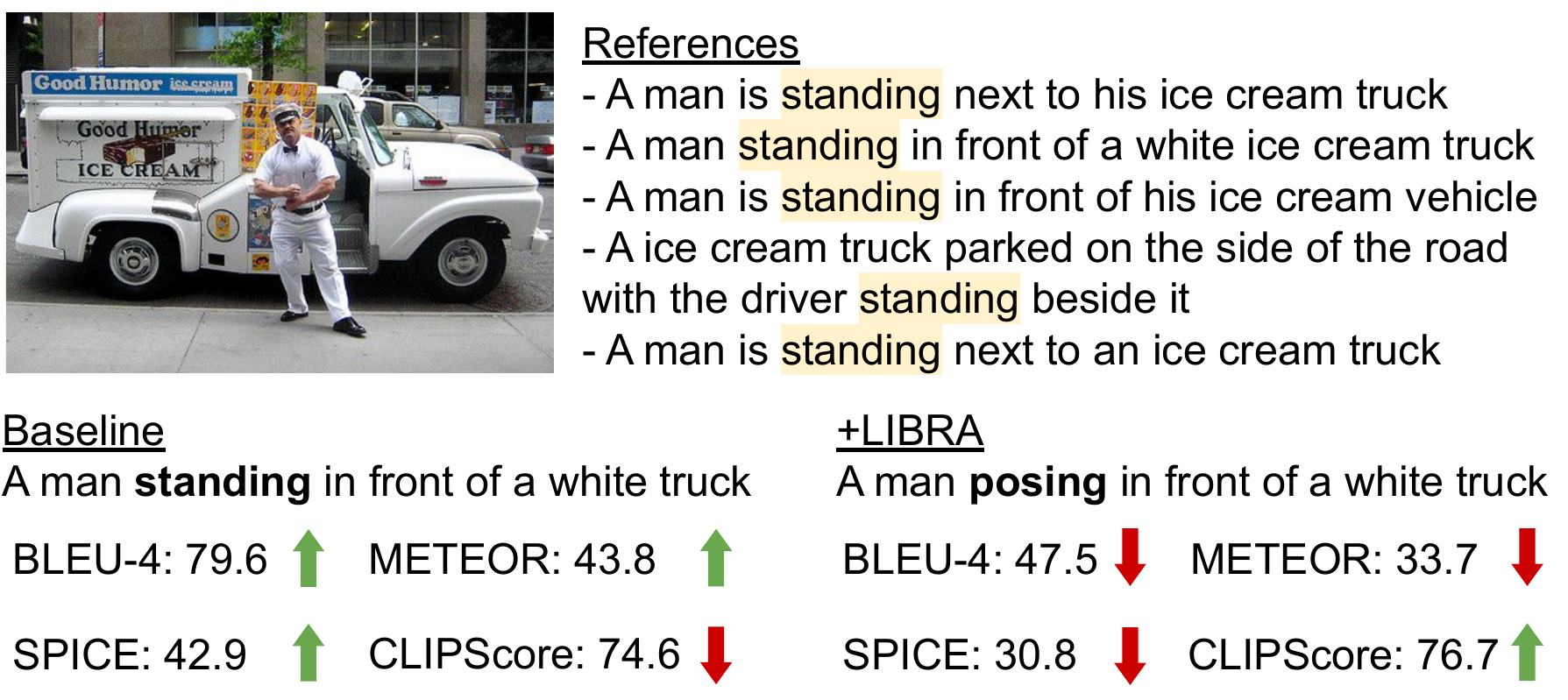}
  \caption{CLIPScore \cite{hessel2021clipscore} vs. reference-based metrics \cite{papineni2002bleu,anderson2016spice,denkowski2014meteor}. References denote the ground-truth captions written by annotators. Bold words in the generated captions mean the difference between baseline and LIBRA. Highlighted words in references denote the words that match the bold word in the baseline. We can see that CLIPScore is more robust against word-changing.}
  \label{fig:clipscore}
\end{figure}

%\begin{table*}[t]
%\renewcommand{\arraystretch}{1.1}
%\setlength{\tabcolsep}{5pt}
%\small
%\centering
%\caption{Comparison with Gender equalizer \cite{burns2018women}. \textcolor{colbest}{Green}/\textcolor{colworst}{red} denotes the bias mitigation method mitigates/amplifies bias from the baselines.}
%\begin{tabularx}{0.77\textwidth}{l r r r r c r r r r r}
%\toprule
%& \multicolumn{3}{c}{Gender bias $\downarrow$} & & \multicolumn{5}{c}{Captioning quality $\uparrow$} \\ \cline{2-4} \cline{6-10}
%Model & LIC & Error & BiasAmp && BLEU-4 & CIDEr & METEOR & SPICE & CLIPScore \\
%\midrule
%\cellcolor{oursrow} NIC+ \cite{vinyals2015show} & \cellcolor{oursrow} 1.4 & \cellcolor{oursrow} 14.6 & \cellcolor{oursrow} 3.34  & \cellcolor{oursrow} & \cellcolor{oursrow} 31.0 & \cellcolor{oursrow} 85.5 & \cellcolor{oursrow} 25.7 & \cellcolor{oursrow} 17.5 & \cellcolor{oursrow} 69.9\\
%\hspace{2pt} +Equalizer \cite{burns2018women}  & \textcolor{colworst}{6.8} & \textcolor{colbest}{7.8} & \textcolor{colbest}{1.58} && 29.8 & 80.6 & 25.0 & 16.8 & 69.9\\
%\hspace{2pt} +LIBRA  & \textcolor{colbest}{0.4} & \textcolor{colbest}{5.1} & \textcolor{colbest}{-2.18} && 32.3 & 90.2 & 26.9 & 18.9 & 72.7\\
%\bottomrule
%\end{tabularx}
%\label{tab:equalizer}
%\end{table*}

We apply LIBRA on top of all the captioning models to evaluate if it mitigates the two types of gender biases. We also report caption evaluation scores based on captioning metrics. Results are shown in Table \ref{tab:main}. We summarize the main observations as follows:

\vspace{5pt}
\noindent
\textbf{LIBRA mitigates \gtoc bias.}
The results on LIC show that applying LIBRA consistently decreases \gtoc bias in all the models. We show some examples of LIBRA mitigating bias in Figure \ref{fig:first} (b). For example, in the second sample from the right, the baseline, UpDn \cite{anderson2018bottom}, produces the incorrect word, \textit{suit}. The word \textit{suit} is skewed toward men, co-occurring with men $82$\% of the time in the MSCOCO training set. LIBRA changes \textit{suit} to \textit{jacket}, mitigating \gtoc bias. 
Besides, in some cases where LIC is negative (\ie, NIC, SAT, FC, Att2in, and ClipCap), the \gtoc bias in the generated captions by LIBRA is less than those of human annotators. In the appendix, we show some examples that LIBRA generates less biased captions than annotators' captions. 

The results of BiasAmp, which LIBRA consistently reduces, show that LIBRA tends to equalize the skewed word-gender co-occurrences. For example, LIBRA mitigates the co-occurrence of the word \textit{little} and women from $91$\% in captions by OSCAR to $60$\%. Results on BiasAmp support the effectiveness of LIBRA regarding the ability to mitigate \gtoc bias.

\vspace{5pt}
\noindent
\textbf{LIBRA mitigates \ctog bias in most models.}
The Error scores show that LIBRA reduces gender misclassification in most models except for OSCAR and GRIT ($3.0 \rightarrow 4.6$ for OSCAR, $3.5 \rightarrow 4.1$ for GRIT). We investigate the error cases when LIBRA is applied to OSCAR and find that gender misclassification of LIBRA is often caused by insufficient evidence to identify a person's gender. For instance, in the top example in Figure \ref{fig:miscls}, the ground-truth gender annotation is \textit{male}, and OSCAR generates \textit{man} although the person is not pictured properly enough to determine gender.\footnote{Previous work \cite{bhargava2019exposing} has shown human annotators possibly annotate gender from context for images without enough cues to judge gender.} This may suggest that OSCAR learns to guess the gender based on the context, in this case, skateboard\footnote{\textit{Skateboard} is highly skewed toward men in the dataset, which co-occurs with men more than $90$\%.} to increase gender classification accuracy. However, this causes \ctog bias for images with a gender-context combination rarely seen in the dataset (\eg, women-surfing). In Figure \ref{fig:miscls} (bottom), OSCAR predicts incorrect gender for the image with a male-biased context.\footnote{\textit{Surfboard} highly co-occur with men in MSCOCO.}
%In Section \ref{sec:limitations}, 
In the appendix, we discuss possible solutions for reducing gender misclassification without relying on the context.

\vspace{5pt}
\noindent
\textbf{LIBRA is good at CLIPScore.}
The results of the captioning metrics show that CLIPScore is better or almost as high as the baselines when applying LIBRA. As CLIPScore is based on an image-caption matching score, we can confirm that LIBRA does not generate less biased sentences by producing irrelevant words to images. This observation verifies that applying LIBRA on top of the captioning models does not hurt the quality of captions. 

\vspace{5pt}
\noindent
\textbf{CLIPScore versus other metrics.}
While LIBRA works well on CLIPScore, the score in the reference-based metrics decreases for some models.  We examine the cause of the inconsistency between CLIPScore and reference-based metrics and find that generating words that reduce bias hurts reference-based metrics. We show an example in Figure \ref{fig:clipscore}. LIBRA changes \textit{standing} to \textit{posing}, which is also a valid description of the image. However, the scores of reference-based metrics substantially drop (\eg, $79.6 \rightarrow 47.5$ in BLEU-4). Human annotators tend to use \textit{posing} for women.\footnote{The co-occurrence of women and \textit{posing} is more than $60$\% of the time in the MSCOCO training set.} Therefore, reference-based metrics value captions that imitate how annotators describe each gender. On the other hand, LIBRA tends to change words skewed toward each gender to more neutral ones, which can be the cause of decreasing scores in the reference-based metrics. %Based on the analysis and the results of CLIPScore, LIBRA does not degrade the quality of the captions compared to the baselines.

\subsection{Comparison with other bias mitigation}

\begin{table}[t]
\renewcommand{\arraystretch}{1.1}
\setlength{\tabcolsep}{5pt}
\small
\centering
\caption{Comparison with Gender equalizer \cite{burns2018women}. \textcolor{colbest}{Green}/\textcolor{colworst}{red} denotes the bias mitigation method mitigates/amplifies bias.}
\begin{tabularx}{0.93\columnwidth}{l r r c r r}
\toprule
& \multicolumn{2}{c}{Gender bias $\downarrow$} & & \multicolumn{2}{c}{Captioning quality $\uparrow$} \\ \cline{2-3} \cline{5-6}
Model & LIC & Error && SPICE & CLIPScore \\
\midrule
\cellcolor{oursrow} NIC+ \cite{vinyals2015show} & \cellcolor{oursrow} 1.4 & \cellcolor{oursrow} 14.6 & \cellcolor{oursrow} & \cellcolor{oursrow} 17.5 & \cellcolor{oursrow} 69.9\\
\hspace{2pt} +Equalizer \cite{burns2018women}  & \textcolor{colworst}{6.8} & \textcolor{colbest}{7.8} && 16.8 & 69.9\\
\hspace{2pt} +LIBRA  & \textcolor{colbest}{0.4} & \textcolor{colbest}{5.1} && 18.9 & 72.7\\
\bottomrule
\end{tabularx}
\label{tab:equalizer}
\end{table}

\begin{figure}[t]
  \centering
  \includegraphics[clip, width=0.98\columnwidth]{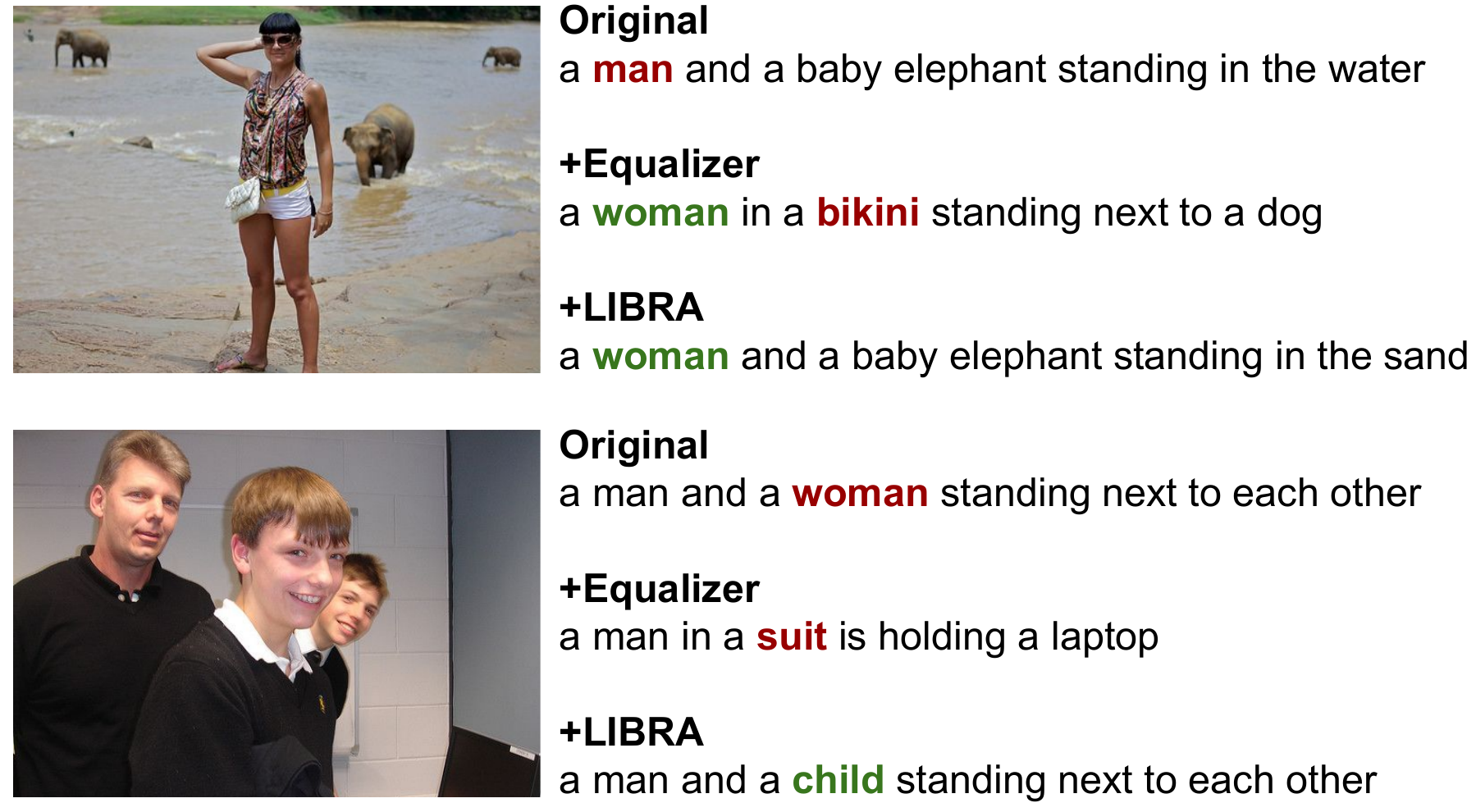}
  \caption{LIBRA vs. Gender equalizer \cite{burns2018women}. }
  \label{fig:equalizer}
\end{figure}

We compare the performance of LIBRA and Gender equalizer \cite{burns2018women} on NIC+ \cite{vinyals2015show}, following the code provided by the authors. The results are shown in Table \ref{tab:equalizer}.  %
%
%As reported in the previous work \cite{hirota2022quantifying,wang2021directional}, Gender equalizer amplifies \gtoc bias ($1.4 \rightarrow 6.8$ in LIC) while mitigating gender misclassification ($14.6 \rightarrow 7.8$ in Error). For example, in the upper sample of Figure \ref{fig:equalizer}, Gender equalizer produces a gender-stereotypical word, \textit{bikini}, whereas it predicts the correct gender. 
%
As reported in previous work \cite{hirota2022quantifying,wang2021directional}, Gender equalizer amplifies \gtoc bias ($1.4 \rightarrow 6.8$ in LIC) while mitigating gender misclassification ($14.6 \rightarrow 7.8$ in Error).
In contrast, LIBRA mitigates \gtoc and \ctog biases, specifically $1.4 \rightarrow 0.4$ in LIC and $14.6 \rightarrow 5.1$ in Error. In the upper sample of Figure \ref{fig:equalizer}, LIBRA predicts the correct gender while not generating gender-stereotypical words. The results of the comparison with Gender equalizer highlight the importance of considering two types of biases for gender bias mitigation.

%\begin{table*}[t]
%\renewcommand{\arraystretch}{1.1}
%\setlength{\tabcolsep}{5pt}
%\small
%\centering
%\caption{Comparison with the caption refinement model}
%\begin{tabularx}{0.92\textwidth}{@{}l r r r r r r c r r r r r@{}}
%\toprule
%& \multicolumn{6}{c}{Gender bias $\downarrow$} & & \multicolumn{4}{c}{Accuracy $\uparrow$} \\ \cline{2-7} \cline{9-13}
%Model & LIC & LIC$_M$  & Error & BA & DBA$_G$ & DBA$_O$ & & BLEU-4	& CIDEr & METEOR & ROUGE-L & CLIP \\
%\midrule
%SAT \cite{xu2015show} & -  & - & - & - & - & - &  & - & - & - & - & -\\
%\hspace{2pt} +ETN \cite{sammani2020show}  & -  & - & - & - & - & - &  & - & - & - & - & -\\
%\hspace{2pt} +LIBRA  & -  & - & - & - & - & - &  & - & - & - & - & -\\
%UpDn \cite{anderson2018bottom} & -  & - & - & - & - & - &  & - & - & - & - & -\\
%\hspace{2pt} +ETN \cite{sammani2020show}  & -  & - & - & - & - & - &  & - & - & - & - & -\\
%\hspace{2pt} +LIBRA  & -  & - & - & - & - & - &  & - & - & - & - & -\\
%OSCAR \cite{li2020oscar} & -  & - & - & - & - & - &  & - & - & - & - & -\\
%\hspace{2pt} +ETN \cite{sammani2020show}  & -  & - & - & - & - & - &  & - & - & - & - & -\\
%\hspace{2pt} +LIBRA  & -  & - & - & - & - & - &  & - & - & - & - & -\\
%\bottomrule
%\end{tabularx}
%\label{tab:genderbias}
%\end{table*}

\subsection{Comparison with image caption editing model}
\label{sec:edit}

\begin{table}[t]
\renewcommand{\arraystretch}{1.1}
\setlength{\tabcolsep}{5pt}
\small
\centering
\caption{Comparison with image caption editing model. Bold numbers represent the best scores in ENT \cite{sammani2020show} or LIBRA.}
\vspace{-5pt}
\begin{tabularx}{0.925\columnwidth}{l r r c r r}
\toprule
& \multicolumn{2}{c}{Gender bias $\downarrow$} & & \multicolumn{2}{c}{Captioning quality $\uparrow$}\\ \cline{2-3}  \cline{5-6}
Model & LIC & Error & & SPICE & CLIPScore  \\
\midrule
%\cellcolor{oursrow} UpDn \cite{anderson2018bottom} & \cellcolor{oursrow} 4.7 & \cellcolor{oursrow} 5.6 & \cellcolor{oursrow} & \cellcolor{oursrow} 22.0 & \cellcolor{oursrow} 73.8\\
%\hspace{2pt} +ENT \cite{sammani2020show}  & 3.9 & 5.6 && 21.3 & 72.5\\
%\hspace{2pt} +LIBRA & \textbf{1.5} & \textbf{4.5} && 22.0 & 74.6\\

\cellcolor{oursrow} OSCAR \cite{li2020oscar} & \cellcolor{oursrow} 2.4 & \cellcolor{oursrow} 3.0 & \cellcolor{oursrow} & \cellcolor{oursrow} 24.0 & \cellcolor{oursrow} 75.8\\
\hspace{2pt} +ENT \cite{sammani2020show} & 5.7 & \textbf{2.8} && 21.9 & 72.8\\
\hspace{2pt} +LIBRA & \textbf{0.3} & 4.6 && 23.2 & 75.7\\
\bottomrule
\end{tabularx}
\label{tab:edit}
\end{table}

We compare LIBRA with a state-of-the-art image caption editing model \cite{sammani2020show} (refer to ENT). Specifically, we apply LIBRA and ENT on top of the various captioning models and evaluate them in terms of bias metrics and captioning metrics. We re-train ENT by using the captions from SAT \cite{xu2015show} for textual features.
%\footnote{In the original paper, the authors use the captions from AoANet \cite{huang2019attention}. We use SAT for training ENT as AoANet is trained on Karpathy split \cite{karpathy2015deep}.} 
The results for OSCAR \cite{li2020oscar} are shown in Table \ref{tab:edit}. The complete results are in the appendix. As for LIC, while LIBRA consistently mitigates \gtoc bias, ENT can amplify the bias in some baselines (SAT, Att2in, OSCAR, ClipCap, GRIT). Regarding Error, LIBRA outperforms in most baselines except for OSCAR and GRIT. From these observations, we conclude that a dedicated framework for addressing gender bias is necessary to mitigate gender bias.

\subsection{Ablations}
\label{sec:ablation}

We conduct ablation studies to analyze the influence of different settings of LIBRA. Here, we show the results when applying LIBRA to UpDn \cite{anderson2018bottom} and OSCAR \cite{li2020oscar}. The complete results of all the baselines are in the appendix. 
%\alert{We conduct ablation studies to analyze the influence of different settings of LIBRA. Here, we show the results when applying LIBRA to OSCAR \cite{li2020oscar}. The complete results of all the baselines are in the appendix. }

\vspace{-12pt}
\paragraph{Combinations of synthetic data}

\begin{table}[t]
\renewcommand{\arraystretch}{1.1}
\setlength{\tabcolsep}{5pt}
\small
\centering
\caption{Comparison of data used for training DCG. Bold numbers denote the best scores among the types of synthetic datasets.}
\vspace{-5pt}
\begin{tabularx}{0.925\columnwidth}{l c c c c r r}
\toprule
& \multicolumn{3}{c}{Synthesis method} & & \multicolumn{2}{c}{Gender bias $\downarrow$}\\ \cline{2-4}  \cline{6-7}
Model & Swap & T5 & Merged & & LIC & Error  \\
\midrule

\cellcolor{oursrow} UpDn \cite{anderson2018bottom} & \cellcolor{oursrow} - & \cellcolor{oursrow} - & \cellcolor{oursrow} - & \cellcolor{oursrow} & \cellcolor{oursrow} 4.7 & \cellcolor{oursrow} 5.6\\
\hspace{2pt} +LIBRA  & \Checkmark & \Checkmark  & - && 2.3 & 6.2\\
\hspace{2pt} +LIBRA   & - & \Checkmark & \Checkmark  && 1.5 & \textbf{4.5}\\
\hspace{2pt} +LIBRA   & \Checkmark & \Checkmark & \Checkmark  && \textbf{1.1} & 5.2\\

\cellcolor{oursrow} OSCAR \cite{li2020oscar} & \cellcolor{oursrow} - & \cellcolor{oursrow} - & \cellcolor{oursrow} - & \cellcolor{oursrow} & \cellcolor{oursrow} 2.4 & \cellcolor{oursrow} 3.0\\
\hspace{2pt} +LIBRA  & \Checkmark & \Checkmark  & - && \textbf{-0.8} & 6.8\\
\hspace{2pt} +LIBRA   & - & \Checkmark & \Checkmark  && 0.3 & \textbf{4.6}\\
\hspace{2pt} +LIBRA   & \Checkmark & \Checkmark & \Checkmark  && 0 & 5.0\\
\bottomrule
\end{tabularx}
\label{tab:data-ablation}
\end{table}

We compare the performance of the different dataset combinations for training DCG in Table \ref{tab:dataset}. 
The results are shown in Table \ref{tab:data-ablation}. The Error score of T5-generation and Merged is consistently the best among the combinations. As for LIC, the results are not as consistent, but still DCG trained on all types of combinations decreases the score. We chose T5-generation and Merged as it well balances LIC and Error.

\vspace{-12pt}
\paragraph{Synthetic data evaluation}

\begin{table}[t]
\renewcommand{\arraystretch}{1.1}
\setlength{\tabcolsep}{5pt}
\small
\centering
\caption{Comparison with random perturbation. Rand.~pert. denotes DCG trained on data with random perturbation. Bold numbers denote the best scores in the DCG trained on either biased captions from BCS or captions with random perturbation.}
\vspace{-5pt}
\begin{tabularx}{0.925\columnwidth}{l r r c r r}
\toprule
& \multicolumn{2}{c}{Gender bias $\downarrow$} & & \multicolumn{2}{c}{Captioning quality $\uparrow$}\\ \cline{2-3}  \cline{5-6}
Model & LIC & Error & & SPICE & CLIPScore  \\
\midrule
\cellcolor{oursrow} UpDn \cite{anderson2018bottom} & \cellcolor{oursrow} 4.7 & \cellcolor{oursrow} 5.6 & \cellcolor{oursrow} & \cellcolor{oursrow} 22.0 & \cellcolor{oursrow} 73.8\\
\hspace{2pt} +Rand.~pert.  & 2.2 & 5.9 && 21.8 & 74.4\\
\hspace{2pt} +LIBRA & \textbf{1.5} & \textbf{4.5} && 22.0 & 74.6\\
\cellcolor{oursrow} OSCAR \cite{li2020oscar} & \cellcolor{oursrow} 2.4 & \cellcolor{oursrow} 3.0 & \cellcolor{oursrow} & \cellcolor{oursrow} 24.0 & \cellcolor{oursrow} 75.8\\
\hspace{2pt} +Rand.~pert.  & 2.0 & 5.6 && 22.9 & 75.4\\
\hspace{2pt} +LIBRA & \textbf{0.3} & \textbf{4.6} && 23.2 & 75.7\\
\bottomrule
\end{tabularx}
\label{tab:random}
\end{table}

To demonstrate the effectiveness of BCS, we compare LIBRA and DCG trained on captions with random perturbation, which does not necessarily increase gender bias. In order to synthesize such captions, we randomly mask $15$ percent of the tokens in the original captions in $\mathcal{D}_\text{g}$ and generate words by T5, but without using any filters in Section \ref{sec:data_aug}. When selecting masked tokens, we allow choosing gender words so that T5 can randomly change the gender. As a result, the synthesized captions contain incorrect words, which are not necessarily due to gender bias. 
We show the results in Table \ref{tab:random}. Using biased samples from BCS to train DCG consistently produces the best results in LIC and Error. From this, we conclude that BCS, which intentionally synthesizes captions with gender biases, contributes to mitigating gender biases.

\section{Conclusion}
\label{sec:conclusion}

\modelname\footnote{This work is partly supported by JST CREST Grant No. JPMJCR20D3, JST FOREST Grant No. JPMJFR216O, JSPS KAKENHI No. JP22K12091, and Grant-in-Aid for Scientific Research (A).} is a model-agnostic framework to mitigate both \ctog and \gtoc biases in captioning models. We experimentally showed that LIBRA mitigates gender bias in multiple captioning models, correcting gender misclassification caused by context and changing to less gender-stereotypical words. To do this, LIBRA synthesizes biased captions using a language model and filtering for intentionally increasing gender biases. Interestingly, the results showed these synthetic captions are a good proxy of gender-biased captions from various captioning models and facilitate model-agnostic bias mitigation. As future work, we will use LIBRA to mitigate other types of bias, such as age or skin-tone, which requires specific annotations, such as the ones in concurrent work \cite{garciauncurated}, and mechanisms to identify each type of bias.

% \alert{\paragraph{Acknowledgments} This work is partly supported by JST CREST Grant No. JPMJCR20D3, JST FOREST Grant No. JPMJFR216O, JSPS KAKENHI No. JP22K12091, and Grant-in-Aid for Scientific Research (A).}

%%%%%%%%% REFERENCES
{\small
\bibliographystyle{ieee_fullname}
\bibliography{egbib}
}

\end{document}

\maketitle

%%%%%%%%% BODY TEXT
This supplementary material includes:

\begin{itemize}[topsep=0pt,itemsep=-1ex,partopsep=1ex,parsep=1ex]
    \item Details of BCS (Appendix~\ref{sec:bcs}).
    \item Details of bias metrics (Appendix~\ref{sec:bias-metrics}).
    \item Additional experiments (Appendix~\ref{sec:addexps}).
    \item More visual examples (Appendix~\ref{sec:examples}).
    \item List of gender-related words (Appendix~\ref{sec:list-words}).
    \item Limitations (Appendix~\ref{sec:limitations}).
    \item Potential negative impact (Appendix~\ref{sec:impact}).
    %\item Code (external files).
\end{itemize}

\section{Details of BCS}
\label{sec:bcs}
In this section, we provide the details for BCS. 

\subsection{Training gender classifier}
The gender classifier $f_\text{g}$ is trained on $\mathcal{D}_\text{g}$. Specifically, following \cite{hirota2022quantifying}, we split the captions in $\mathcal{D}_\text{g}$ into a balanced split with an equal number of samples per female/male, having $66,526$ captions. We use BERT-base \cite{devlin2018bert} with two fully-connected layers with Leaky ReLU as $f_\text{g}$ and finetune it on the balanced split. The learning rate is $1 \times 10^{-5}$, and the training is conducted on $5$ epochs. 

\subsection{Finetuning T5}
Following the masked language model in \cite{devlin2018bert}, we finetune T5 on captions in $\mathcal{D}$. Specifically, we mask $10$\% of the tokens in the original caption $y$. Given the masked caption and the positions of masked tokens $\mathcal{M} = \{m_1, \dots ,m_{|\mathcal{M}|}\}$, T5 predicts the probability of masked tokens by $\prod_{m \in \mathcal{M}} p(y_m|y_{\backslash \mathcal{M}})$, where $y_{\backslash \mathcal{M}}$ denotes all words in an input caption $y$ except for masked tokens $\{y_m\}$. The sample-wise loss is defined as: 
\begin{equation}
    \mathcal{L}_{mlm} = -\text{log} \prod_{m \in \mathcal{M}} p(y_m|y_{\backslash \mathcal{M}})
\end{equation}
where $p(y_m|y_{\backslash \mathcal{M}})$ is the output probability of masked token $y_m$ given $y_{\backslash \mathcal{M}}$ from T5.  

\subsection{Details of T5 masked word generation}
To remove trivial modifications, we avoid generating synonyms of the masked tokens by using WordNet \cite{miller1995wordnet}. When selecting masked tokens, we chose nouns/verbs/adjectives/adverbs based on POS tagging with NLTK \cite{loper2002nltk}. 

We apply a filter to remove unnatural captions, called an authenticity filter. The authenticity filter uses a classifier that predicts whether an input sentence is synthetic or authentic. 
Specifically, we train classifier $f_\text{a}$ with $\mathcal{D}_\text{T5,g} \cup \mathcal{D}_g$ to predict whether $y$ is from $\mathcal{D}_\text{T5,g}$ or $\mathcal{D}_g$. Let $b \in \{\text{syn}, \text{auth}\}$, prediction $\hat{b}$ is given by:
\begin{equation}
    \hat{b} = f_\text{a}(y) = \text{argmax}_b \ p(B=b | y)
\end{equation}
where $p(B=b | y)$ is a confidence score that an input caption is $b$. Thus, if $p(B=\text{authy} | y)$ is close to $1$, $y$ is likely to be authentic. We use this classifier to filter less natural captions, \ie,
\begin{equation}
    F_\text{AF}(\mathcal{D}_\text{T5,g}) = \{y \in \mathcal{D}_\text{T5,g}|p(B=\text{auth}|y) > \alpha)\},
\end{equation}
where $\alpha$ is a predefined threshold. We set $\alpha = 0.3$ and use the same classifier as $f_\text{g}$ for $f_\text{a}$. 
%We show in the appendix that this authenticity filter can detect sentences that contain grammatical errors and unnatural-sounding descriptions. 

\subsection{Examples of gender/authenticity filter}

\begin{figure}[t]
  \centering
  \includegraphics[clip, width=1.0\columnwidth]{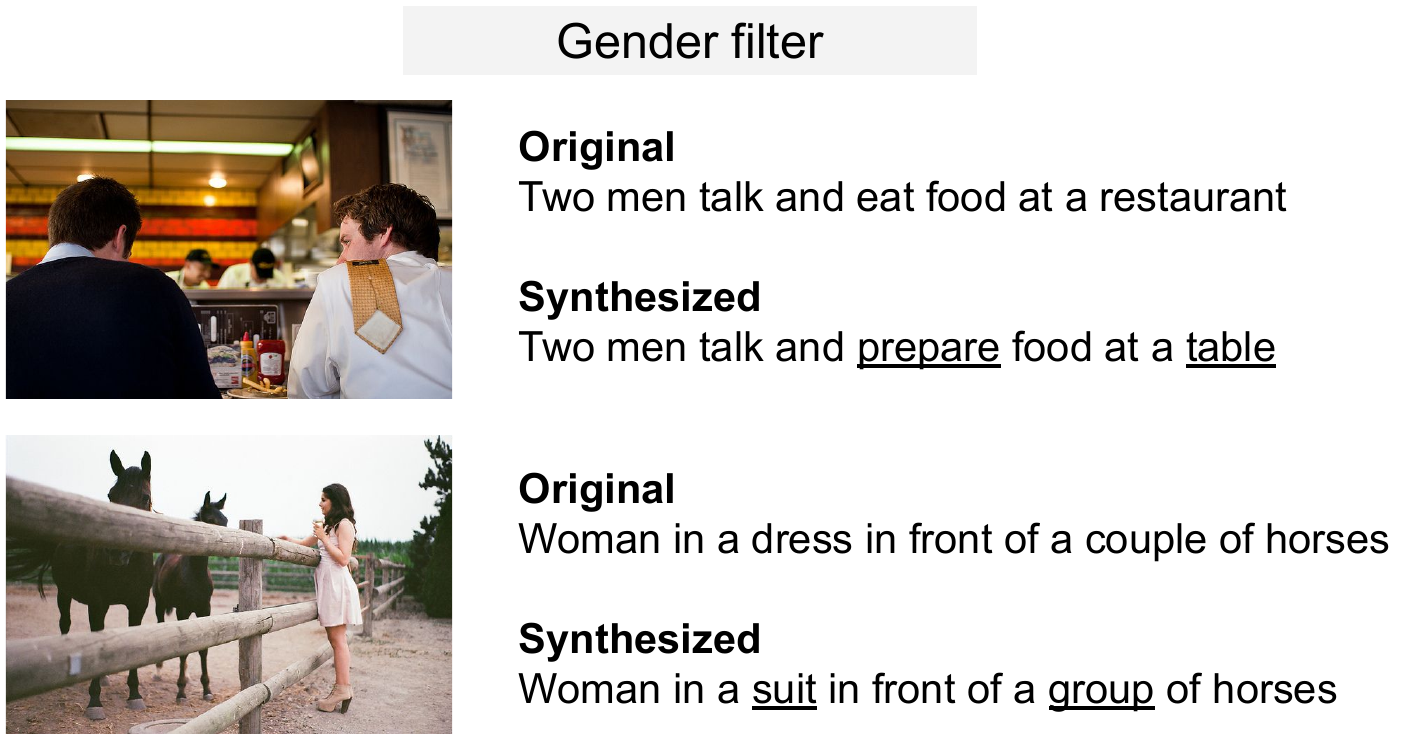}
  \caption{Synthesized captions that are removed by the gender filter.}
  \label{fig:gender-filter}
\end{figure}

\begin{table*}[t]
  \caption{Synthesized captions that are passed or removed by the authenticity filter}
  %\vspace{-8pt}
  \label{tab:auth-filter}
  \begin{center}
  \begin{tabular}{l|l|l}
    \midrule
    Original & Passed &  Removed\\
    \hline
    Woman is sitting near a red train & Woman is sitting near a passenger train & Woman sits sitting near a red train\\ 
    A man wearing glasses, suit, and tie & A man wearing sunglasses, hat, and tie  & A man wearing glasses, glasses, and tie\\
    A man fixing the inside of a toilet &  A man fixing the inside of a kitchen  &  A man holding the inside of a toilet\\
    Women are playing a video game & Women are playing a baseball game & Women are playing a video show\\
    \midrule
  \end{tabular}
  \end{center}
\end{table*}

\begin{table*}[t]
\renewcommand{\arraystretch}{1.1}
\setlength{\tabcolsep}{5pt}
\small
\centering
\caption{Comparison with image caption editing models. Bold numbers represent the best scores in ENT \cite{sammani2020show} or LIBRA.} 
\begin{tabularx}{0.71\textwidth}{l r r c r r r r r}
\toprule
& \multicolumn{2}{c}{Gender bias $\downarrow$} & & \multicolumn{5}{c}{Captioning quality $\uparrow$}\\ \cline{2-3}  \cline{5-9}
Model & LIC & Error && BLEU-4 & CIDEr & METEOR & SPICE & CLIPScore  \\
\midrule
\cellcolor{oursrow} NIC \cite{vinyals2015show} & \cellcolor{oursrow} 0.5 & \cellcolor{oursrow} 23.6 & \cellcolor{oursrow} & \cellcolor{oursrow} 21.9 & \cellcolor{oursrow} 58.3 & \cellcolor{oursrow} 21.6 & \cellcolor{oursrow} 13.4 & \cellcolor{oursrow} 65.2\\
    \hspace{2pt} +ENT \cite{sammani2020show} & \textbf{-0.3} & 22.5 && 25.8 & 67.7 & 22.5 & 14.3 & 65.3\\
\hspace{2pt} +LIBRA & \textbf{-0.3} & \textbf{5.7}  && 24.6 & 72.0 & 24.2 & 16.5 & 71.7\\
\cellcolor{oursrow} SAT \cite{xu2015show} & \cellcolor{oursrow} -0.3 & \cellcolor{oursrow} 9.1 & \cellcolor{oursrow} &  \cellcolor{oursrow} 34.5 & \cellcolor{oursrow} 94.6 & \cellcolor{oursrow} 27.3 &\cellcolor{oursrow} 19.2 & \cellcolor{oursrow} 72.1\\
\hspace{2pt} +ENT \cite{sammani2020show} & 1.6 & 9.9 && 35.3 & 96.3 & 27.3 & 19.2 & 71.1\\
\hspace{2pt} +LIBRA & \textbf{-1.4} & \textbf{3.9}  && 34.6 & 95.9 & 27.8 & 20.0 & 73.6\\
\cellcolor{oursrow} FC \cite{rennie2017self} & \cellcolor{oursrow} 2.9 & \cellcolor{oursrow} 10.3 & \cellcolor{oursrow} & \cellcolor{oursrow} 32.2 & \cellcolor{oursrow} 94.2 & \cellcolor{oursrow} 26.1 &\cellcolor{oursrow} 18.3 & \cellcolor{oursrow} 70.0\\
\hspace{2pt} +ENT \cite{sammani2020show} & 1.7 & 10.3 && 32.9 & 92.0 & 26.2 & 18.2 & 69.2\\
\hspace{2pt} +LIBRA & \textbf{-0.2} & \textbf{4.3}  && 32.8 & 95.9 & 27.3 & 19.7 & 72.9\\
\cellcolor{oursrow} Att2in \cite{rennie2017self} & \cellcolor{oursrow} 1.1 & \cellcolor{oursrow} 5.4 & \cellcolor{oursrow} & \cellcolor{oursrow} 36.7 & \cellcolor{oursrow} 102.8 & \cellcolor{oursrow} 28.4 &\cellcolor{oursrow} 20.2 & \cellcolor{oursrow} 72.6\\
\hspace{2pt} +ENT \cite{sammani2020show} & 2.8 & 5.3 && 37.4 & 103.2 & 28.4 & 20.3 & 71.6\\
\hspace{2pt} +LIBRA & \textbf{-0.3} & \textbf{4.6}  && 35.9 & 101.7 & 28.5 & 20.6 & 73.8\\
\cellcolor{oursrow} UpDn \cite{anderson2018bottom} & \cellcolor{oursrow} 4.7 & \cellcolor{oursrow} 5.6 & \cellcolor{oursrow} & \cellcolor{oursrow} 39.4 & \cellcolor{oursrow} 115.1 & \cellcolor{oursrow} 29.8 &\cellcolor{oursrow} 22.0 & \cellcolor{oursrow} 73.8\\
\hspace{2pt} +ENT \cite{sammani2020show} & 3.9 & 5.6 && 39.6 & 110.7 & 29.4 & 21.3 & 72.5\\
\hspace{2pt} +LIBRA & \textbf{1.5} & \textbf{4.5}  && 37.7 & 110.1 & 29.6 & 22.0 & 74.6\\
\cellcolor{oursrow} Transformer \cite{vaswani2017attention} & \cellcolor{oursrow} 5.4 & \cellcolor{oursrow} 6.9 & \cellcolor{oursrow} & \cellcolor{oursrow} 35.0 & \cellcolor{oursrow} 101.5 & \cellcolor{oursrow} 28.9 &\cellcolor{oursrow} 21.1 & \cellcolor{oursrow} 75.3\\
\hspace{2pt} +ENT \cite{sammani2020show} & 4.4 & 6.8 && 38.6 & 107.1 & 28.9 & 20.8 & 72.9\\
\hspace{2pt} +LIBRA & \textbf{2.3} & \textbf{5.0}  && 33.9 & 98.7 & 28.6 & 20.9 & 75.7\\
\cellcolor{oursrow} OSCAR \cite{li2020oscar} & \cellcolor{oursrow} 2.4 & \cellcolor{oursrow} 3.0 & \cellcolor{oursrow} & \cellcolor{oursrow} 39.4 & \cellcolor{oursrow} 119.8 & \cellcolor{oursrow} 32.1 &\cellcolor{oursrow} 24.0 & \cellcolor{oursrow} 75.8\\
\hspace{2pt} +ENT \cite{sammani2020show} & 5.7 & \textbf{2.8} && 41.4 & 113.0 & 30.2 & 21.9 & 72.8\\
\hspace{2pt} +LIBRA & \textbf{0.3} & 4.6  && 37.2 & 113.1 & 31.1 & 23.2 & 75.7\\
\cellcolor{oursrow} ClipCap \cite{mokady2021clipcap} & \cellcolor{oursrow} 1.1 & \cellcolor{oursrow} 5.6 & \cellcolor{oursrow} & \cellcolor{oursrow} 34.8 & \cellcolor{oursrow} 103.7 & \cellcolor{oursrow} 29.6 &\cellcolor{oursrow} 21.5 & \cellcolor{oursrow} 76.6\\
\hspace{2pt} +ENT \cite{sammani2020show} & 3.6 & 5.1 && 37.4 & 101.7 & 28.4 & 20.1 & 73.0\\
\hspace{2pt} +LIBRA & \textbf{-1.5} & \textbf{4.5}  && 33.8 & 100.6 & 29.3 & 21.4 & 76.0\\
\cellcolor{oursrow} GRIT \cite{nguyen2022grit} & \cellcolor{oursrow} 3.1 & \cellcolor{oursrow} 3.5 & \cellcolor{oursrow} & \cellcolor{oursrow} 42.9 & \cellcolor{oursrow} 123.3 & \cellcolor{oursrow} 31.5 &\cellcolor{oursrow} 23.4 & \cellcolor{oursrow} 76.2\\
\hspace{2pt} +ENT \cite{sammani2020show} & 5.2 & \textbf{3.7} && 42.8 & 120.3 & 30.8 & 22.7 & 74.0\\
\hspace{2pt} +LIBRA & \textbf{0.7} & 4.1  && 40.5 & 116.8 & 30.6 & 22.6 & 75.9\\
\bottomrule
\end{tabularx}
\label{tab:edit}
\end{table*}

\begin{table}[t]
\renewcommand{\arraystretch}{1.1}
\setlength{\tabcolsep}{5pt}
\small
\centering
\caption{Comparison with DCG without masking input captions. Bold numbers denote the best scores in the DCG with/without masking.}
\begin{tabularx}{0.98\columnwidth}{l r r c r r}
\toprule
& \multicolumn{2}{c}{Gender bias $\downarrow$} & & \multicolumn{2}{c}{Accuracy $\uparrow$}\\ \cline{2-3}  \cline{5-6}
Model & LIC & Error & & SPICE & CLIPScore  \\
\midrule
\cellcolor{oursrow} NIC \cite{vinyals2015show} & \cellcolor{oursrow} 0.5 & \cellcolor{oursrow} 23.6 & \cellcolor{oursrow} & \cellcolor{oursrow} 13.4 & \cellcolor{oursrow} 65.2\\
\hspace{2pt} +DCG w/o mask & \textbf{-2.1} & 5.9 && 15.7 & 70.5\\
\hspace{2pt} +LIBRA & -0.3 & \textbf{5.7}  && 16.5 & 71.7\\
\cellcolor{oursrow} SAT \cite{xu2015show} & \cellcolor{oursrow} -0.3 & \cellcolor{oursrow} 9.1 & \cellcolor{oursrow} & \cellcolor{oursrow} 19.2 & \cellcolor{oursrow} 72.1\\
\hspace{2pt} +DCG w/o mask & -1.3 & 4.0 && 19.8 & 72.8\\
\hspace{2pt} +LIBRA & \textbf{-1.4} & \textbf{3.9}  && 20.0 & 73.6\\
\cellcolor{oursrow} FC \cite{rennie2017self} & \cellcolor{oursrow} 2.9 & \cellcolor{oursrow} 10.3 & \cellcolor{oursrow} & \cellcolor{oursrow} 18.3 & \cellcolor{oursrow} 70.0\\
\hspace{2pt} +DCG w/o mask & 0.5 & 4.4 && 19.6 & 72.0\\
\hspace{2pt} +LIBRA & \textbf{-0.2} & \textbf{4.3}  && 19.7 & 72.9\\
\cellcolor{oursrow} Att2in \cite{rennie2017self} & \cellcolor{oursrow} 1.1 & \cellcolor{oursrow} 5.4 & \cellcolor{oursrow} & \cellcolor{oursrow} 20.2 & \cellcolor{oursrow} 72.6\\
\hspace{2pt} +DCG w/o mask & 0.7 & \textbf{4.6} && 20.6 & 73.0\\
\hspace{2pt} +LIBRA & \textbf{-0.3} & \textbf{4.6}  && 20.6 & 73.8\\
\cellcolor{oursrow} UpDn \cite{anderson2018bottom} & \cellcolor{oursrow} 4.7 & \cellcolor{oursrow} 5.6 & \cellcolor{oursrow} & \cellcolor{oursrow} 22.0 & \cellcolor{oursrow} 73.8\\
\hspace{2pt} +DCG w/o mask & 1.9 & 4.8 && 21.9 & 73.8\\
\hspace{2pt} +LIBRA & \textbf{1.5} & \textbf{4.5}  && 22.0 & 74.6\\
\cellcolor{oursrow} Transformer \cite{vaswani2017attention} & \cellcolor{oursrow} 5.4 & \cellcolor{oursrow} 6.9 & \cellcolor{oursrow} & \cellcolor{oursrow} 21.1 & \cellcolor{oursrow} 75.3\\
\hspace{2pt} +DCG w/o mask & 4.4 & 5.6 && 20.9 & 74.9\\
\hspace{2pt} +LIBRA & \textbf{2.3} & \textbf{5.0}  && 20.9 & 75.7\\
\cellcolor{oursrow} OSCAR \cite{li2020oscar} & \cellcolor{oursrow} 2.4 & \cellcolor{oursrow} 3.0 & \cellcolor{oursrow} & \cellcolor{oursrow} 24.0 & \cellcolor{oursrow} 75.8\\
\hspace{2pt} +DCG w/o mask & 1.9 & 4.7 && 23.4 & 75.8\\
\hspace{2pt} +LIBRA & \textbf{0.3} & \textbf{4.6}  && 23.2 & 75.7\\
\cellcolor{oursrow} ClipCap \cite{mokady2021clipcap} & \cellcolor{oursrow} 1.1 & \cellcolor{oursrow} 5.6 & \cellcolor{oursrow} & \cellcolor{oursrow} 21.5 & \cellcolor{oursrow} 76.6\\
\hspace{2pt} +DCG w/o mask & 0.5 & 4.7 && 21.4 & 76.2\\
\hspace{2pt} +LIBRA & \textbf{-1.5} & \textbf{4.5}  && 21.4 & 76.0\\
\cellcolor{oursrow} GRIT \cite{nguyen2022grit} & \cellcolor{oursrow} 3.1 & \cellcolor{oursrow} 3.5 & \cellcolor{oursrow} & \cellcolor{oursrow} 23.4 & \cellcolor{oursrow} 76.2\\
\hspace{2pt} +DCG w/o mask & 1.8 & 4.3 && 22.8 & 75.3\\
\hspace{2pt} +LIBRA & \textbf{0.7} & \textbf{4.1}  && 22.6 & 75.9\\
\bottomrule
\end{tabularx}
\label{tab:masking}
\end{table}

\begin{table}[t]
\renewcommand{\arraystretch}{1.1}
\setlength{\tabcolsep}{5pt}
\small
\centering
\caption{Comparison of data used for training DCG. Bold numbers denote the best scores among the types of synthetic datasets.}
\begin{tabularx}{1.0\columnwidth}{l c c c c r r}
\toprule
& \multicolumn{3}{c}{Synthesis method} & & \multicolumn{2}{c}{Gender bias $\downarrow$}\\ \cline{2-4}  \cline{6-7}
Model & Swap & T5 & Merged & & LIC & Error  \\
\midrule

\cellcolor{oursrow}  NIC \cite{vinyals2015show} & \cellcolor{oursrow} - & \cellcolor{oursrow} - & \cellcolor{oursrow} - & \cellcolor{oursrow} & \cellcolor{oursrow} 0.5 & \cellcolor{oursrow} 23.6\\
\hspace{2pt} +LIBRA  & \Checkmark & \Checkmark  & - && -0.1 & 7.5\\
\hspace{2pt} +LIBRA   & - & \Checkmark & \Checkmark  && \textbf{-0.3} & \textbf{5.7}\\
\hspace{2pt} +LIBRA   & \Checkmark & \Checkmark & \Checkmark  && -0.2 & 6.2\\

\cellcolor{oursrow}  SAT \cite{xu2015show} & \cellcolor{oursrow} - & \cellcolor{oursrow} - & \cellcolor{oursrow} - & \cellcolor{oursrow} & \cellcolor{oursrow} -0.3 & \cellcolor{oursrow} 9.1\\
\hspace{2pt} +LIBRA  & \Checkmark & \Checkmark  & - && -2.0 & 6.2\\
\hspace{2pt} +LIBRA   & - & \Checkmark & \Checkmark  && -1.4 & \textbf{3.9}\\
\hspace{2pt} +LIBRA   & \Checkmark & \Checkmark & \Checkmark  && \textbf{-2.3} & 4.8\\

\cellcolor{oursrow}  FC \cite{rennie2017self} & \cellcolor{oursrow} - & \cellcolor{oursrow} - & \cellcolor{oursrow} - & \cellcolor{oursrow} & \cellcolor{oursrow} 2.9 & \cellcolor{oursrow} 10.3\\
\hspace{2pt} +LIBRA  & \Checkmark & \Checkmark  & - && 0.5 & 6.5\\
\hspace{2pt} +LIBRA   & - & \Checkmark & \Checkmark  && -0.2 & \textbf{4.3}\\
\hspace{2pt} +LIBRA   & \Checkmark & \Checkmark & \Checkmark  && \textbf{-0.9} & 5.0\\

\cellcolor{oursrow} Att2in \cite{rennie2017self} & \cellcolor{oursrow} - & \cellcolor{oursrow} - & \cellcolor{oursrow} - & \cellcolor{oursrow} & \cellcolor{oursrow} 1.1 & \cellcolor{oursrow} 5.4\\
\hspace{2pt} +LIBRA  & \Checkmark & \Checkmark  & - && 2.0 & 6.7\\
\hspace{2pt} +LIBRA   & - & \Checkmark & \Checkmark  && -0.3 & \textbf{4.6}\\
\hspace{2pt} +LIBRA   & \Checkmark & \Checkmark & \Checkmark  && \textbf{-1.2} & 5.5\\

\cellcolor{oursrow} UpDn \cite{anderson2018bottom} & \cellcolor{oursrow} - & \cellcolor{oursrow} - & \cellcolor{oursrow} - & \cellcolor{oursrow} & \cellcolor{oursrow} 4.7 & \cellcolor{oursrow} 5.6\\
\hspace{2pt} +LIBRA  & \Checkmark & \Checkmark  & - && 2.3 & 6.2\\
\hspace{2pt} +LIBRA   & - & \Checkmark & \Checkmark  && 1.5 & \textbf{4.5}\\
\hspace{2pt} +LIBRA   & \Checkmark & \Checkmark & \Checkmark  && \textbf{1.1} & 5.2\\

\cellcolor{oursrow} Transformer \cite{vaswani2017attention} & \cellcolor{oursrow} - & \cellcolor{oursrow} - & \cellcolor{oursrow} - & \cellcolor{oursrow} & \cellcolor{oursrow} 5.4 & \cellcolor{oursrow} 6.9\\
\hspace{2pt} +LIBRA  & \Checkmark & \Checkmark  & - && \textbf{1.5} & 6.9\\
\hspace{2pt} +LIBRA   & - & \Checkmark & \Checkmark  && 2.3 & \textbf{5.0}\\
\hspace{2pt} +LIBRA   & \Checkmark & \Checkmark & \Checkmark  && 2.6 & 5.8\\

\cellcolor{oursrow} OSCAR \cite{li2020oscar} & \cellcolor{oursrow} - & \cellcolor{oursrow} - & \cellcolor{oursrow} - & \cellcolor{oursrow} & \cellcolor{oursrow} 2.4 & \cellcolor{oursrow} 3.0\\
\hspace{2pt} +LIBRA  & \Checkmark & \Checkmark  & - && \textbf{-0.8} & 6.8\\
\hspace{2pt} +LIBRA   & - & \Checkmark & \Checkmark  && 0.3 & \textbf{4.6}\\
\hspace{2pt} +LIBRA   & \Checkmark & \Checkmark & \Checkmark  && 0 & 5.0\\

\cellcolor{oursrow} ClipCap \cite{mokady2021clipcap} & \cellcolor{oursrow} - & \cellcolor{oursrow} - & \cellcolor{oursrow} - & \cellcolor{oursrow} & \cellcolor{oursrow} 1.1 & \cellcolor{oursrow} 5.6\\
\hspace{2pt} +LIBRA  & \Checkmark & \Checkmark  & - && -1.3 & 6.8\\
\hspace{2pt} +LIBRA   & - & \Checkmark & \Checkmark  && -1.5 & \textbf{4.5}\\
\hspace{2pt} +LIBRA   & \Checkmark & \Checkmark & \Checkmark  && \textbf{-1.7} & 5.3\\

\cellcolor{oursrow} GRIT \cite{nguyen2022grit} & \cellcolor{oursrow} - & \cellcolor{oursrow} - & \cellcolor{oursrow} - & \cellcolor{oursrow} & \cellcolor{oursrow} 3.1 & \cellcolor{oursrow} 3.5\\
\hspace{2pt} +LIBRA  & \Checkmark & \Checkmark  & - && \textbf{-0.8} & 6.3\\
\hspace{2pt} +LIBRA   & - & \Checkmark & \Checkmark  && 0.7 & \textbf{4.1}\\
\hspace{2pt} +LIBRA   & \Checkmark & \Checkmark & \Checkmark  && 0 & 4.8\\
\bottomrule
\end{tabularx}
\label{tab:data-ablation}
\end{table}

In Figure \ref{fig:gender-filter}, we show some synthesized captions that are filtered out by the gender filter. The removed captions do not increase gender bias with respect to the original captions. For instance, in the bottom example, the word \textit{dress} which is skewed toward women is replaced with \textit{suit} which is skewed toward men. Thus the synthesized caption reduces gender bias compared to the original caption, and it is filtered out by the gender filter.  

In Table \ref{tab:auth-filter}, we show some synthesized captions that are passed or removed by the authenticity filter. The examples show that the authenticity filter removes unnatural-sounding or grammatically incorrect captions.

\section{Details of bias metrics}
\label{sec:bias-metrics}

%\paragraph{LIC}
%We follow the settings of the original implementations by the authors \cite{hirota2022quantifying}. The classifier is BERT-base \cite{devlin2018bert}, and we finetune the classifier for $5$ epochs.  

\paragraph{BiasAmp}
As for BiasAmp, we also follow the settings presented in \cite{hirota2022quantifying}.  
To compute gender-word co-occurrences, we use the top $1,000$ frequent words in $\mathcal{D}$. Specifically, we select nouns/verbs/adjectives/adverbs in the top $1,000$ words. Following \cite{zhao2017men}, we use words that are strongly related to humans by removing words that do not appear more than $100$ times with women/men words.

\section{Additional experiments}
\label{sec:addexps}

\subsection{Comparison with image caption editing model}

We compare LIBRA with a state-of-the-art image caption editing model \cite{sammani2020show} (refer to ENT). Specifically, we apply LIBRA and ENT on top of the various captioning models and evaluate them in terms of bias metrics and captioning metrics. We re-train ENT by using the captions from SAT \cite{xu2015show} for textual features.\footnote{In the original paper, the authors use the captions from AoANet \cite{huang2019attention}. We use SAT for training ENT as AoANet is trained on Karpathy split \cite{karpathy2015deep}.} The results are shown in Table \ref{tab:edit}. As for LIC, while LIBRA consistently mitigates \gtoc bias, ENT can amplify the bias in some baselines (SAT, Att2in, OSCAR, ClipCap, GRIT). Regarding Error, LIBRA outperforms in most baselines except for OSCAR and GRIT. From these observations, we conclude that a dedicated framework for addressing gender bias is necessary to mitigate gender bias.

\subsection{Analysis of masking}

We evaluate the effectiveness of masking input captions in DCG. Specifically, we compare LIBRA with DCG whose input captions are not masked (\ie, $\eta = 0$). 
The results are shown in Table \ref{tab:masking}. We can see that masking the input captions of DCG consistently improves the scores on bias metrics, which contributes to mitigating two types of biases.

\subsection{Complete results of ablations}

Here, we show the complete results of the ablations in the main paper. 

\paragraph{Combinations of synthetic data}

The complete results of all the baselines are shown in Table \ref{tab:data-ablation}. As in the analysis of the main paper, the results of LIC are not as consistent while DCG trained on all types of combinations mitigate \gtoc bias. Regarding Error, DCG trained on T5-generation and Merged has the best results.

\paragraph{Synthetic data evaluation}

\begin{table}[t]
\renewcommand{\arraystretch}{1.1}
\setlength{\tabcolsep}{5pt}
\small
\centering
\caption{Comparison with random perturbation. Rand.~pert. denotes DCG trained on data with random perturbation. Bold numbers denote the best scores in the DCG trained on either biased captions from BCS or captions with random perturbation.}
\begin{tabularx}{1.0\columnwidth}{l r r c r r}
\toprule
& \multicolumn{2}{c}{Gender bias $\downarrow$} & & \multicolumn{2}{c}{Accuracy $\uparrow$}\\ \cline{2-3}  \cline{5-6}
Model & LIC & Error & & SPICE & CLIPScore  \\
\midrule
\cellcolor{oursrow} NIC \cite{vinyals2015show} & \cellcolor{oursrow} 0.5 & \cellcolor{oursrow} 23.6 & \cellcolor{oursrow} & \cellcolor{oursrow} 13.4 & \cellcolor{oursrow} 65.2\\
\hspace{2pt} +Rand.~pert. mask & 0.7 & 7.7 && 16.4 & 71.5\\
\hspace{2pt} +LIBRA & \textbf{-0.3} & \textbf{5.7}  && 16.5 & 71.7\\
\cellcolor{oursrow} SAT \cite{xu2015show} & \cellcolor{oursrow} -0.3 & \cellcolor{oursrow} 9.1 & \cellcolor{oursrow} & \cellcolor{oursrow} 19.2 & \cellcolor{oursrow} 72.1\\
\hspace{2pt} +Rand.~pert. & \textbf{-1.5} & 6.5 && 19.9 & 73.4\\
\hspace{2pt} +LIBRA & -1.4 & \textbf{3.9}  && 20.0 & 73.6\\
\cellcolor{oursrow} FC \cite{rennie2017self} & \cellcolor{oursrow} 2.9 & \cellcolor{oursrow} 10.3 & \cellcolor{oursrow} & \cellcolor{oursrow} 18.3 & \cellcolor{oursrow} 70.0\\
\hspace{2pt} +Rand.~pert. & 0.2 & 6.6 && 19.8 & 72.7\\
\hspace{2pt} +LIBRA & \textbf{-0.2} & \textbf{4.3}  && 19.7 & 72.9\\
\cellcolor{oursrow} Att2in \cite{rennie2017self} & \cellcolor{oursrow} 1.1 & \cellcolor{oursrow} 5.4 & \cellcolor{oursrow} & \cellcolor{oursrow} 20.2 & \cellcolor{oursrow} 72.6\\
\hspace{2pt} +Rand.~pert. & \textbf{-0.8} & 5.9 && 20.4 & 73.7\\
\hspace{2pt} +LIBRA & -0.3 & \textbf{4.6}  && 20.6 & 73.8\\
\cellcolor{oursrow} UpDn \cite{anderson2018bottom} & \cellcolor{oursrow} 4.7 & \cellcolor{oursrow} 5.6 & \cellcolor{oursrow} & \cellcolor{oursrow} 22.0 & \cellcolor{oursrow} 73.8\\
\hspace{2pt} +Rand.~pert. & 2.2 & 5.9 && 21.8 & 74.4\\
\hspace{2pt} +LIBRA & \textbf{1.5} & \textbf{4.5}  && 22.0 & 74.6\\
\cellcolor{oursrow} Transformer \cite{vaswani2017attention} & \cellcolor{oursrow} 5.4 & \cellcolor{oursrow} 6.9 & \cellcolor{oursrow} & \cellcolor{oursrow} 21.1 & \cellcolor{oursrow} 75.3\\
\hspace{2pt} +Rand.~pert. & 3.6 & 6.2 && 20.7 & 75.4\\
\hspace{2pt} +LIBRA & \textbf{2.3} & \textbf{5.0}  && 20.9 & 75.7\\
\cellcolor{oursrow} OSCAR \cite{li2020oscar} & \cellcolor{oursrow} 2.4 & \cellcolor{oursrow} 3.0 & \cellcolor{oursrow} & \cellcolor{oursrow} 24.0 & \cellcolor{oursrow} 75.8\\
\hspace{2pt} +Rand.~pert. & 2.0 & 5.6 && 22.9 & 75.4\\
\hspace{2pt} +LIBRA & \textbf{0.3} & \textbf{4.6}  && 23.2 & 75.7\\
\cellcolor{oursrow} ClipCap \cite{mokady2021clipcap} & \cellcolor{oursrow} 1.1 & \cellcolor{oursrow} 5.6 & \cellcolor{oursrow} & \cellcolor{oursrow} 21.5 & \cellcolor{oursrow} 76.6\\
\hspace{2pt} +Rand.~pert. & 0.5 & 5.9 && 21.2 & 75.8\\
\hspace{2pt} +LIBRA & \textbf{-1.5} & \textbf{4.5}  && 21.4 & 76.0\\
\cellcolor{oursrow} GRIT \cite{nguyen2022grit} & \cellcolor{oursrow} 3.1 & \cellcolor{oursrow} 3.5 & \cellcolor{oursrow} & \cellcolor{oursrow} 23.4 & \cellcolor{oursrow} 76.2\\
\hspace{2pt} +Rand.~pert. & 1.8 & 5.6 && 22.4 & 75.8\\
\hspace{2pt} +LIBRA & \textbf{0.7} & \textbf{4.1}  && 22.6 & 75.9\\
\bottomrule
\end{tabularx}
\label{tab:pert}
\end{table}

Table \ref{tab:pert} shows the results of the comparison with random perturbation. This extended table also shows that biased samples from BCS to train DCG produces the best results in LIC and Error in most baselines, which shows the effectiveness of BCS in mitigating gender bias.

\section{More visual examples}
\label{sec:examples}

\paragraph{Bias mitigation by LIBRA}

\begin{figure*}[t]
  \centering
  \includegraphics[clip, width=1.0\textwidth]{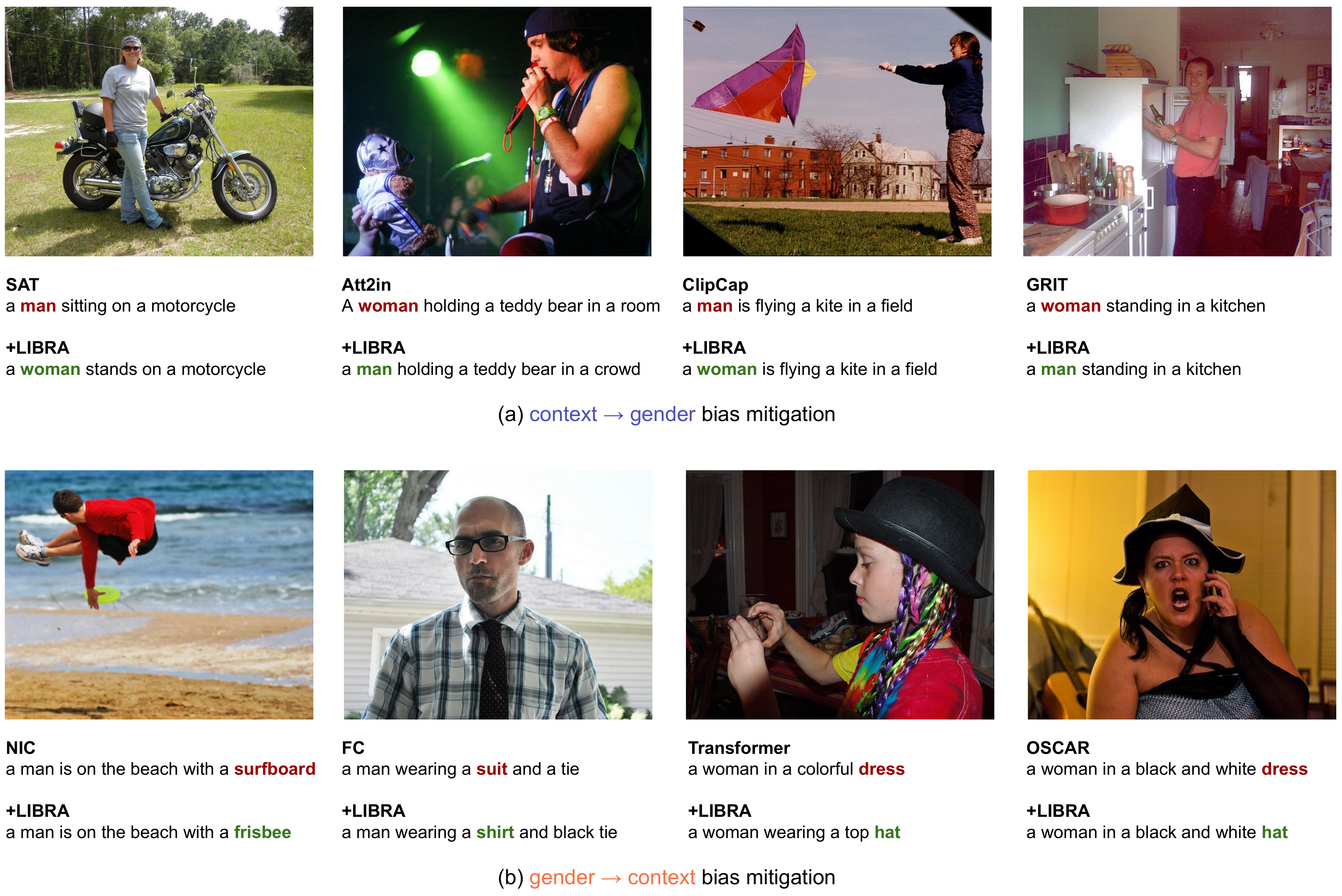}
  \caption{Generated captions by the baseline captioning models and LIBRA. We show the baseline suffers from \ctog/\gtoc biases, predicting incorrect gender or incorrect word. Our proposed framework successfully modifies those incorrect words.}
  \label{fig:first-supp}
\end{figure*}

Figure \ref{fig:first-supp} shows the additional examples that LIBRA mitigates \ctog or \gtoc bias. For instance, in the left example of (a), the word \textit{motorcycle} highly co-occurs with men in the MSCOCO training set,\footnote{Co-occurrence of \textit{Motorcycle} and men is about $2.7$ times the co-occurrence of \textit{Motorcycle} and women.} and the baseline predicts the incorrect gender \textit{man} probably due to \ctog bias. Applying LIBRA on top of the baseline results in mitigating that bias by predicting the correct gender.

\paragraph{Synthesized captions by BCS}

\begin{figure*}[t]
  \centering
  \includegraphics[clip, width=1.0\textwidth]{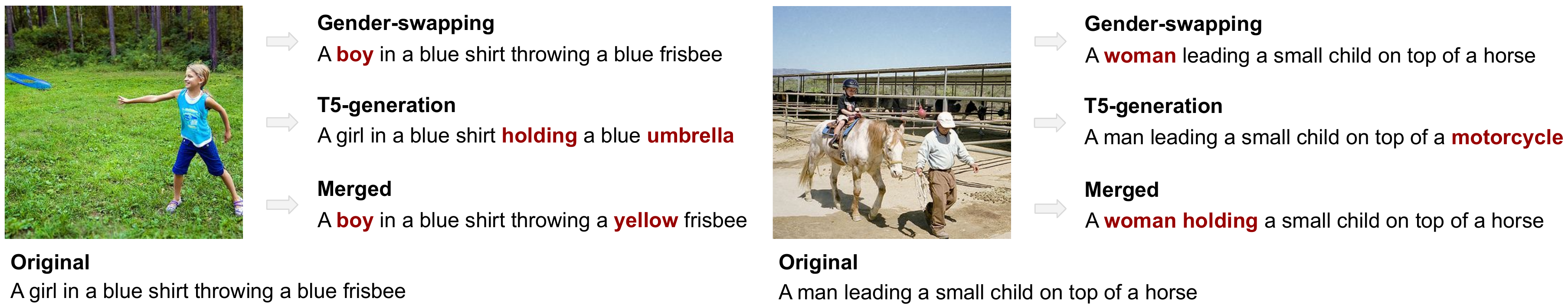}
  \caption{Biased captions synthesized by BCS.}
  \label{fig:bcs-supp}
\end{figure*}

In Figure \ref{fig:bcs-supp}, we show some additional examples of the synthesized captions by BCS. The synthesized captions contain \ctog or/and \gtoc biases.

\paragraph{LIBRA vs.~human captions}

\begin{figure*}[t]
  \centering
  \includegraphics[clip, width=1.0\textwidth]{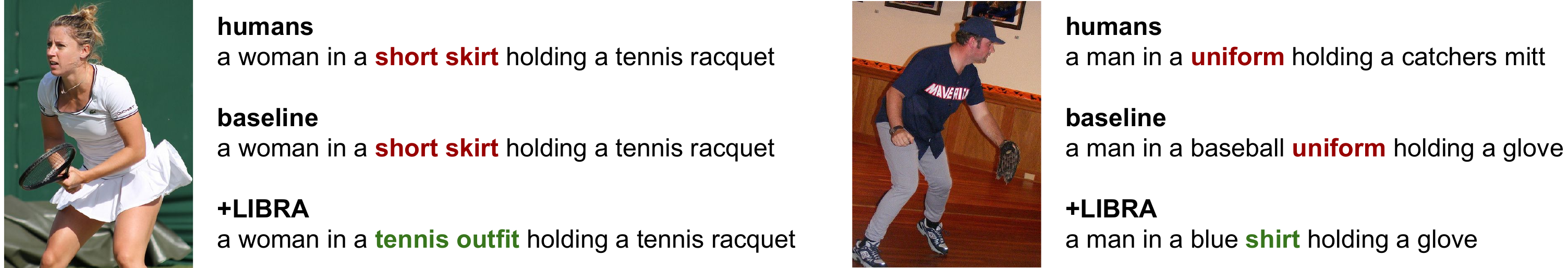}
  \caption{Comparison of captions from human annotators, baseline, and LIBRA.}
  \label{fig:human}
\end{figure*}

The experimental results in the main paper show that LIBRA generates less biased captions than human annotations, resulting in negative LIC scores. \Cref{fig:human} shows some visual examples that LIBRA generates more neutral words than human captions. For instance, in the left sample, both human and baseline captions contain \textit{short skirt}, which is women's stereotypical words while LIBRA uses more neutral words \textit{tennis outfit}.

\begin{figure*}[t]
  \centering
  \includegraphics[clip, width=1.0\textwidth]{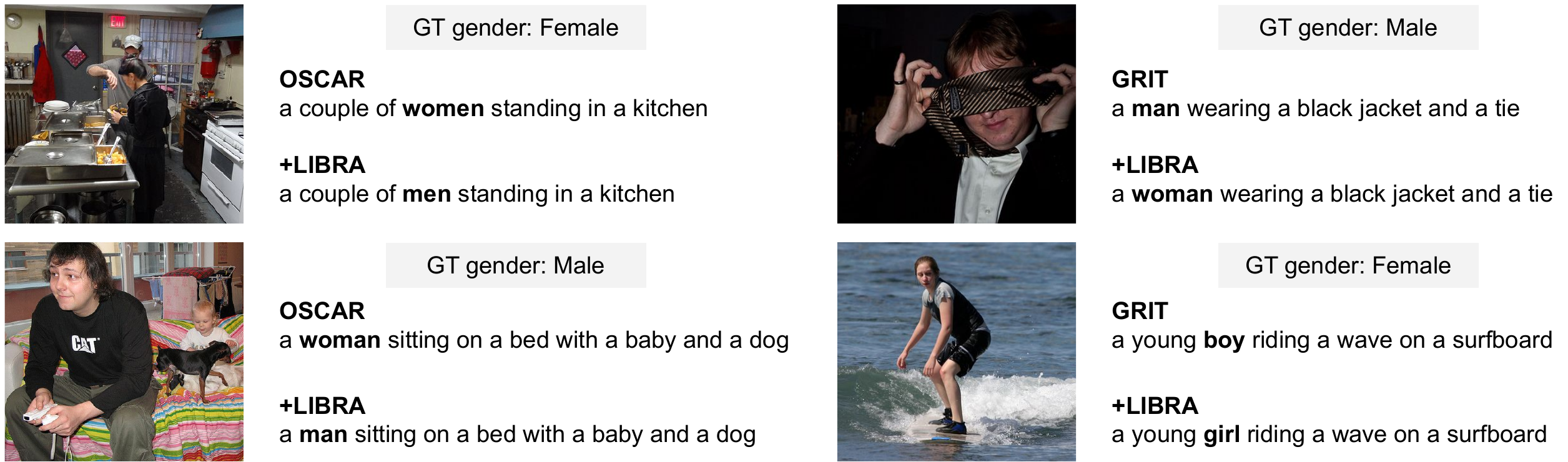}
  \caption{Gender misclassification of LIBRA (Top). Gender misclassification of OSCAR \cite{li2020oscar} or GRIT \cite{nguyen2022grit} (Bottom). GT gender denotes ground-truth gender annotation in \cite{zhao2021captionbias}.}
  \label{fig:miscls-supp}
\end{figure*}

\begin{figure*}[t]
  \centering
  \includegraphics[clip, width=0.99\textwidth]{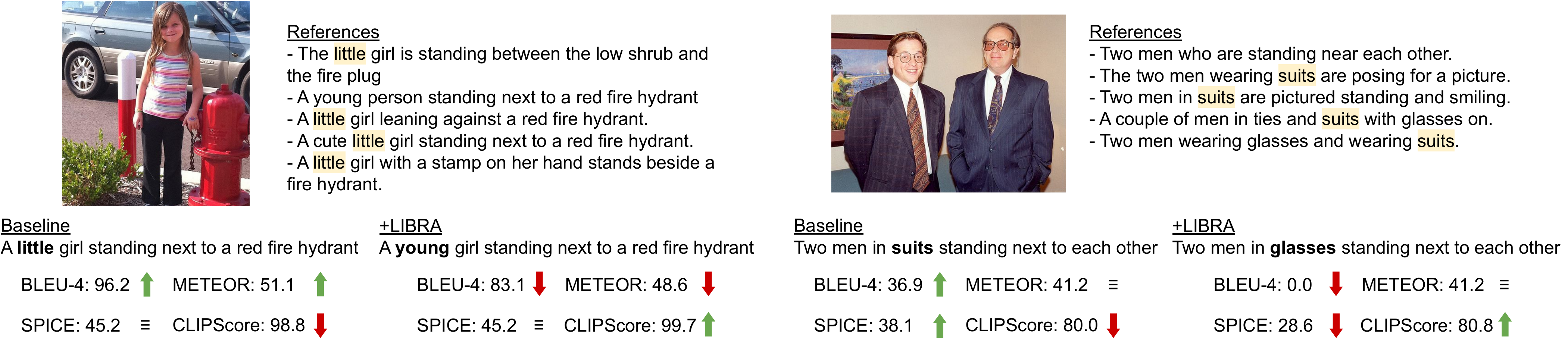}
  \caption{CLIPScore \cite{hessel2021clipscore} vs. reference-based metrics \cite{papineni2002bleu,anderson2016spice,denkowski2014meteor}. References denote the ground-truth captions written by annotators. Bold words in the generated captions mean the difference between baseline and LIBRA. Highlighted words in references denote the words that match the bold word in the baseline. We can see that CLIPScore is more robust against word changing.}
  \label{fig:clipscore-supp}
\end{figure*}

\paragraph{Error cases of LIBRA vs.~state-of-the-art models}

In Figure \ref{fig:miscls-supp}, we show the additional examples of the error cases of LIBRA and the state-of-the-art models, OSCAR \cite{li2020oscar} and GRIT \cite{nguyen2022grit}. As in the explanation in the main paper, state-of-the-art models can guess gender from the context when there is no clear evidence to identify gender, which leads to amplify \ctog bias.

\paragraph{CLIPScore vs.~reference-based metrics}

In Figure \ref{fig:clipscore-supp}, we show the additional examples that LIBRA hurts reference-based metrics by generating words that reduce bias whereas LIBRA does not hurt CLIPScore \cite{hessel2021clipscore}. For instance, in the left example, the word \textit{little} is skewed toward women in the training set, and LIBRA changed it to \textit{young} which is the less biased word.\footnote{The co-occurrence of women and \textit{little} is more than $70$\% of the time in the MSCOCO training set, while \textit{young} is balanced between the gender.} Both captions correctly describe the image, but LIBRA degrades the scores for the reference-based metrics as human annotators tend to use \textit{little} for women. On the other hand, CLIPScore is more robust against such word-changing.

\section{List of gender words}
\label{sec:list-words}

The gender words that consist of women and men words are as below: 

\noindent
\textcolor{Melon}{woman}, \textcolor{Melon}{female}, \textcolor{Melon}{lady}, \textcolor{Melon}{mother}, \textcolor{Melon}{girl}, \textcolor{Melon}{aunt}, \textcolor{Melon}{wife}, \textcolor{Melon}{actress}, \textcolor{Melon}{princess}, \textcolor{Melon}{waitress},  \textcolor{Melon}{daughter}, \textcolor{Melon}{sister}, \textcolor{Melon}{queen}, \textcolor{Melon}{chairwoman}, \textcolor{Melon}{policewoman}, \textcolor{Melon}{girlfriend}, \textcolor{Melon}{pregnant}, \textcolor{Melon}{daughter}, \textcolor{Melon}{she}, \textcolor{Melon}{her}, \textcolor{Melon}{hers}, \textcolor{Melon}{herself},
\textcolor{olive}{\textit{man}}, \textcolor{olive}{\textit{male}}, \textcolor{olive}{\textit{father}}, \textcolor{olive}{\textit{gentleman}}, \textcolor{olive}{\textit{boy}}, \textcolor{olive}{\textit{uncle}}, \textcolor{olive}{\textit{husband}}, \textcolor{olive}{\textit{actor}}, \textcolor{olive}{\textit{prince}}, \textcolor{olive}{\textit{waiter}}, \textcolor{olive}{\textit{son}}, \textcolor{olive}{\textit{brother}}, \textcolor{olive}{\textit{guy}}, \textcolor{olive}{\textit{emperor}}, \textcolor{olive}{\textit{dude}}, \textcolor{olive}{\textit{cowboy}}, \textcolor{olive}{boyfriend}, \textcolor{olive}{chairman}, \textcolor{olive}{policeman}, \textcolor{olive}{\textit{he}}, \textcolor{olive}{\textit{his}}, \textcolor{olive}{\textit{him}}, \textcolor{olive}{\textit{himself}} and their plurals. \textcolor{Melon}{Orange}/\textcolor{olive}{\textit{Olive}} denote women / men words, respectively.

\section{Limitations}
\label{sec:limitations}
While LIBRA shows superior performance in mitigating gender bias, it also presents some limitations. 

\vspace{-12pt}
\noindent
\paragraph{Attributes other than gender} 
Gender tends to be described in captions. However, other types of societal biases such as racial bias may not appear as explicitly mentioned in the text and tend to be more subtle, for which our bias mitigation method may not work properly.
% can be problematic when they are described with discriminatory expressions. Thus, more inclusive mitigation methods that work for various attributes are desirable.

\vspace{-12pt}
\noindent
\paragraph{Error for measuring \ctog bias}
Even though Error can measure one of the aspects of \ctog bias where models make an incorrect prediction of gender based on the context, it does not directly evaluate this bias as it can also occur when predictions are correct but based on the context. Thus, a metric dedicated to \ctog bias would be more insightful.

\vspace{-12pt}
\noindent
\paragraph{Predicting gender-neutral words}
In Section 5.1 in the main paper, we showed that gender misclassification by LIBRA is likely to be caused by the deficient clues to judge gender. A possible solution to mitigate such misclassification without exploiting contextual cues would be to force the model to predict gender-neutral words such as \textit{person} when there is not enough information to judge gender. We leave this extension as future work.  

\section{Potential negative impact}
\label{sec:impact}
While LIBRA mitigates gender bias in the bias metrics, it does not ensure that LIBRA completely removes bias. In other words, even though LIBRA works on the bias metrics, the captioning models can still be biased. Thus, a potential negative impact of the use of LIBRA to mitigate gender bias is that the users of LIBRA may become overly confident that LIBRA eliminates gender bias and overlook the problem of gender bias in captioning models. We should carefully consider gender bias in image captioning as it can also exist in aspects not measured by existing metrics.

%%%%%%%%% REFERENCES
{\small
\bibliographystyle{ieee_fullname}
\bibliography{egbib}
}